\documentclass[runningheads]{llncs}
\usepackage{cite}

\usepackage{amsmath,amssymb,amsfonts}

\usepackage{algorithmicx}
\usepackage{algorithm}
\usepackage{algpseudocode}
\usepackage{graphicx}
\usepackage{textcomp}
\usepackage{xcolor}
\usepackage[nolist]{acronym}
\usepackage{amssymb,amsmath,amsthm}
\usepackage[hidelinks]{hyperref}
\usepackage{array,multirow}
\usepackage[utf8]{inputenc}
\usepackage[english]{babel}
\usepackage{subcaption}
\usepackage[nolist]{acronym}
\usepackage{placeins}
\usepackage{graphicx}
\usepackage[load-configurations=version-1]{siunitx} 
\usepackage[title]{appendix}
\def\*#1{\mathbf{#1}}
\def\b#1{\boldsymbol{#1}}

\newcolumntype{H}{>{\setbox0=\hbox\bgroup}c<{\egroup}@{}}
\newcommand{\norm}[1]{\left\lVert#1\right\rVert_F^2}
\begin{acronym}[PCTKVM] 
    \acro{PCVM}[PCVM]{Probabilistic Classification Vector Machine}
    \acro{PCTKVM}[PCTKVM]{Probabilistic Classification Transfer Kernel Vector Machine}
    \acro{BT}[BT]{Basis-Transfer}
    \acro{TKL}[TKL]{Transfer Kernel Learning}
    \acro{TCA}[TCA]{Transfer Component Analysis}
    \acro{SVM}[SVM]{Support Vector Machine}
    \acro{JDA}[JDA]{Joint Distribution Adaptation}
    \acro{SVD}[SVD]{Singular Value Decomposition}
    \acro{EVD}{Eigenvalue Decomposition}
    \acro{RKHS}[RKHS]{Reproducing Kernel Hilbert Space}
    \acro{PCA}[PCA]{Principal Component Analysis}
    \acro{NBT}[NBT]{Nyström Basis Transfer}
    \acro{SA}[SA]{Subspace Alignment}
    \acro{SVM}[SVM]{Support Vector Machine}
    \acro{NTVM}[NTVM]{Nyström Transfer Vector Machine}
    \acro{SURF}[SURF]{Speeded Up Robust Features Extraction}
    \acro{NSO}[NSO]{Nysröm Subspace Override}
\end{acronym}
\DeclareMathOperator*{\argmin}{\arg\!\min}
\def\BibTeX{{\rm B\kern-.05em{\sc i\kern-.025em b}\kern-.08em
    T\kern-.1667em\lower.7ex\hbox{E}\kern-.125emX}}

\begin{document}

\title{Low-Rank Subspace Override for Unsupervised Domain Adaptation}
\titlerunning{Low-Rank Subspace Override}

\author{Christoph Raab$^1$ \and
Frank-Michael Schleif$^1$}
\authorrunning{C. Raab and F. Schleif}
%
\institute{University for Applied Sciences W{\"u}rzburg-Schweinfurt, Sanderheinrichsleitenweg 20, W{\"u}rzburg, Germany, \email{\{christoph.raab,frank-michael.schleif\}@fhws.de}\\}

\maketitle

\begin{abstract}
Current supervised learning models cannot generalize well across domain boundaries, which is a known problem in many applications, such as robotics or visual classification. Domain adaptation methods are used to improve these generalization properties. However, these techniques suffer either from being restricted to a particular task, such as visual adaptation, require a lot of computational time and data, which is not always guaranteed, have complex parameterization, or expensive optimization procedures. In this work, we present an approach that requires only a well-chosen snapshot of data to find a single domain invariant subspace. The subspace is calculated in closed form and overrides domain structures, which makes it fast and stable in parameterization. By employing low-rank techniques, we emphasize on descriptive characteristics of data.
The presented idea is evaluated on various domain adaptation tasks such as text and image classification against state of the art domain adaptation approaches and achieves remarkable performance across all tasks.
\keywords{Transfer Learning  \and Domain-Adaptation \and Single Value Decomposition \and Nyström approximation \and Subspace Override}
\end{abstract}

\section{Introduction}\label{sec:introduction}
Supervised learning and, in particular, classification is an essential task in machine learning with a broad range of applications.
The obtained models are used to predict the labels of unseen test samples.
A basic assumption in supervised learning is that the underlying domain or distribution is not changing between training and test samples.
If the domain is changing from one task to a related but different task, one would like to reuse the available learning model.
Domain differences are quite common in real-world scenarios and, eventually, lead to substantial performance drops~\cite{Weiss2016}.

\begin{figure}[htbp]
    \centering
    \begin{subfigure}[b]{.15\textwidth}
        \includegraphics[width=1\textwidth]{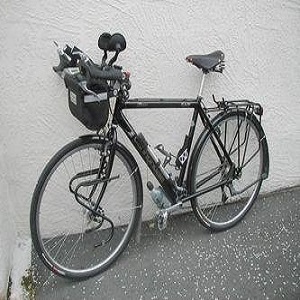}
    \end{subfigure}
    \begin{subfigure}[b]{.15\textwidth}
        \includegraphics[width=1\textwidth]{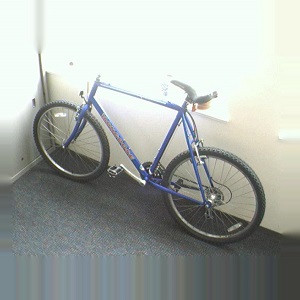}
           \end{subfigure}
    \begin{subfigure}[b]{.15\textwidth}
        \includegraphics[width=1\textwidth]{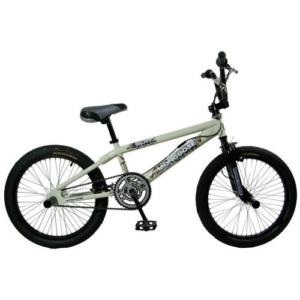}
    \end{subfigure}
    \caption{Objects from different domains~\cite{Gong2012}\label{fig:example_figures}}
\end{figure}

In image classification, a domain adaptation problem exists when the source and target data come from different cameras, as shown in Fig.~\ref{fig:example_figures}. The domain adaptation problem occurs due to different camera characteristics between training and evaluation since cameras have different rendering and focus properties.
More formally, let $\*{X}_s=\{\*x_s^i\}_{i=1}^m \in \mathbb{R}^d$ be $m$ source data samples in a $d$-dimensional feature space from the source domain distribution $p(x_s)$ with labels $Y_s = \{y_s^i\}_{i=1}^m \in \mathcal{Y}=\{1,2,..,C\}$ and let $\*{X}_t=\{\*x_t^j\}_{j=1}^n \in \mathbb{R}^d$ be $n$ target samples from the target domain distribution $p(x_t)$ with labels $Y_t = \{y_t^j\}_{j=1}^n \in \mathcal{Y}$. Traditional machine learning assumes similar distributions, i.e.\ $p(x_s) \sim p(x_t)$, but domain adaptation assumes different distributions, i.e.\ $p(x_s) \neq p(x_t)$.

Various domain adaptation techniques have already been proposed, following different strategies and improving the prediction performance of underlying classification algorithms in test scenarios~\cite{Weiss2016,5288526}. State of the art domain adaptation approaches~\cite{8100030,meda2018,Mahadevan19,Ghifary2017,Long2015} require a large number of source or target samples, which is indeed a disadvantage of many domain adaptation approaches and is not guaranteed in restricted environments where labeling is expensive~\cite{Weiss2016}.
In this work, we show that only a well-chosen subset of samples is necessary to approximate domain structures.

Despite the popularity of kernelized subspace adaptations~\cite{meda2018,Long2013a,8100030} or manifold embeddings~\cite{Gong2012,meda2018,Pan2011,Ghifary2017} for domain alignment, it was shown in~\cite{Blitzer2011,Fernando2013a} that least-squares approaches are at least competitive to more complicated settings, where domain differences are explicitly solved using least-squares to find a common subspace. Solutions to least-square problems are intuitive and theoretically justified. However, if both domains do not lie in a common subspace, this technique fails to transfer knowledge effectively~\cite{Shao2018}. We address this problem and evaluate a domain invariant subspace, where both domains are explicitly part of the target subspace, which neglects the mentioned drawback.

The main contribution of this work is to derive a \textit{subspace} closed-form solution of the least-squares domain adaptation problem by finding a suitable domain invariant projection operator called Subspace Override (SO).
The approach constructs a target subspace representation for both domains, which transfers target basis information to source data. We show that a well-chosen  snapshot of the data is sufficient to approximate the domain characteristics by approximating the optimal solution of the least-squares problem. For the first time in domain adaptation, a Nyström approximation is used on \textit{subspace} domain adaptation. The resulting method has a better prediction performance with stable parameterization and is easy to apply. Further, it is the fastest subspace domain adaptation algorithm in terms of computational complexity compared to related approaches, while maintaining its very good performance.

The rest of the paper is organized as follows: We give an overview of related work in Sec.~\ref{sec:relatedwork}. The underlying mathematical concepts are given in Sec.~\ref{sec:preliminaries}. The proposed approach is discussed in Sec.~\ref{sec:bt}, followed by an experimental part in Sec.~\ref{sec:experiments}, addressing the classification performance, computational time and the stability of the approach.
A summary with a discussion of open issues is provided in the conclusion at the end of the paper. \textbf{Source code, including all experiments and plots, is available at \url{https://github.com/ChristophRaab/nso}.}
\section{Related Work}\label{sec:relatedwork}
In general, homogeneous transfer learning~\cite{Weiss2016} or domain adaptation (DA) approaches, distinguish roughly between the following strategies:

The \textit{feature adaptation} techniques~\cite{Weiss2016} are trying to find a common latent subspace for source and target domain to reduce distribution differences, such that the underlying structure of the data is preserved in the subspace.
A baseline approach for feature adaptation is \ac{TCA}~\cite{Pan2011}. \ac{TCA} finds a suitable subspace transformation called transfer components via minimizing the Maximum Mean Discrepancy (MMD) in the Reproducing Kernel Hilbert Space (RKHS). \ac{JDA}~\cite{Long2013a} also considers MMD but incorporates class-dependent distributions. These works considered a subspace projection based on a combined eigendecomposition for both domains, which fails to include domain-specific attributes into the subspace. The Joint Geometrical Subspace Alignment (JGSA)~\cite{8100030} tackled this issue by searching MMD based subspaces for the domains individually.
However, these methods rely on kernels and are not able to explore the full characteristics of the original feature space and are computationally intensive. Proposed work relies on original space and uses only a snapshot of data for computational efficiency.


\textit{Least-Squares (LS) adaptation} is closely related to us, aligning both domains by finding a solution to the LS problem and use this solution as a feature transformation matrix. The transformation directly modifies the data or finds a subspace projection based on the eigenvectors of the domains. \ac{SA}~\cite{Fernando2013a} computes a target subspace representation by direct modification of the correlation matrices of both domains. The Correlation Alignment (CORAL)~\cite{sun2016return} technique transfers second-order statistics of the target domain into whitened source data and project source and target data via principal component analysis (PCA) into the subspace.
The Landmarks Selection-based Subspace Alignment (LSSA)~\cite{Aljundi2015} is a successor of SA and selects only a subset of both domains, which are near to domain borders to align these borders in the subspace explicitly.However, LSSA cannot capture the whole domain characteristic, and in supervised classification problems, the landmark sample is prone to omit class-information. Our work considers a uniform and class-wise sample strategy to capture the whole domain.

The work of Shao et al.~\cite{Shao2014a} proposed that least-squares approaches, as above, are unable for effective adaptation, because the source and target data may lay not in a single subspace.
In this work, we \textit{override} the orthogonal basis of the source domain with the target one. With this, we model the source subspace domain as part of the target subspace, and subspace differences do not exist because both must lie in the same subspace by construction.

The considered domain adaptation methods have approximately a complexity of $\mathcal{O}(n^2)$, where $n$ is the highest number of samples concerning target or source.
All these algorithms require some \textit{unlabeled test data} to be available at training time.
These transfer-solutions cannot be directly used as predictors, but instead, are wrappers for classification algorithms.

\section{Subspace Override}\label{sec:bt}

The task of domain adaptation is to align distribution differences with the goal that underlying statistics will be  similar afterward. As in prior work~\cite{Aljundi2015,7486497,Elhadji-Ille-Gado2018,Fernando2013a,Shao2014a,Xiao2019,Long2015,stvm}, we assume that similar matrices will lead to similar distributions. Hence, we strive for aligning the domain data matrices in a suitable subspace and model the source data to be part of the target data, and therefore it \textit{must} be in the same (single) subspace.

To draw both domains closer together, represented by their respective samples $\*X_s$ and $\*X_t$, consider the following optimization function
\begin{align}\label{eq:transfer_opt_problem}
    \argmin\limits_{\*M} || \*M\*X_s - \*X_t||_F^2,\\
     s.t. \>\>\> \*M\*M^T = \*I.
\end{align}
The goal is to learn $\*M$ to adapt $\*X_s$ to the target domain. Further, we also make sure that the obtained projection operator is an orthogonal basis.
This formulation has two flaws.

First, if sample sizes of source and target are not the same, i.\,e. $m \neq n$, the above formula is invalid. We address the problem by a simple data augmentation strategy. If $m < n$, $\*X_s$ is enriched by sampling new source data from the estimated Gaussian distribution of $\*X_s$ and assign random source labels until $m = n$. If $n < m$, source samples are randomly removed until sample sizes are equal. Hence, from know we assume $m = n$.

Further,~\eqref{eq:transfer_opt_problem} prevents effective domain adaptation, because the transformation $\*M$ may project the data in different spaces~\cite{Shao2014a}. However, if we model $\*M$ to be directly related to the target domain, the projection operator will be domain invariant.
To get this kind of solution for problem~\eqref{eq:transfer_opt_problem}, it must be rewritten that source data is part of the target subspace.

Let us consider the relationship between singular- and eigendecomposition and rewrite the PCA in terms of SVD. Given a rectangular matrix $\*X \in \mathbb{R}^{n\times d}$ we can rewrite the eigendecomposition to
\begin{equation}\label{eq:pca_svd}
    \*X^T\*X= (\*V \b\Sigma^T \*U^T) (\*U \b\Sigma \*V^T) = \*V \b\Sigma^2 \*V^T,
\end{equation}
with $\b\Sigma \in \mathbb{R}^{n\times d}$ as singular values and $\*U \in \mathbb{R}^{n\times n}$ are singular vectors of $\*X$. Further, $\b\Sigma^2 = \b\Sigma^T\b\Sigma \in \mathbb{R}^{d \times d}$ as eigenvalues and $\*V \in \mathbb{R}^{d\times d}$ as eigenvectors of $\*X^T\*X$. A low rank solution and a reduction of dimensionality is integrated into the new data matrix by sorting $\b\Sigma$ and $\*V$ in descending order with respect to $\b\Sigma$ and choose only the biggest $l$ eigenvalues and corresponding eigenvectors
\begin{equation}\label{eq:proof_valid_pca}
    \*X^l = \*X\*V^l = \*U^l \b\Sigma^l {\*V^l}^T \*V^l = \*U^l \b\Sigma^l \in \mathbb{R}^{n\times l},
\end{equation}
with $\*U^l \in \mathbb{R}^{n\times l}$ and $\b\Sigma^l \in \mathbb{R}^{l\times l}$ and $\*V^l \in \mathbb{R}^{d\times l}$. $\*X^l$ is the reduced target matrix and only the most relevant data w.r.t. to variance is kept. In~\eqref{eq:pca_svd} a linear covariance or kernel is used, but non-linear kernels like the RBF kernel could be integrated as well.

With the insights of~\eqref{eq:pca_svd} and~\eqref{eq:proof_valid_pca}, we rewrite the optimization problem in~\eqref{eq:transfer_opt_problem} to  a low-rank subspace version and state the \textbf{main optimization problem}:
\begin{align}\label{eq:transfer_opt_problem_subspace}
     \argmin\limits_{\*M} || \*M\*U_s^l \b\Sigma_s^l - \*U_t^l \b\Sigma_t^l||_F^2,\\
     s.t. \>\>\> \*M\*M^T = \*I.
\end{align}

Based on domain relatedness and standardization techniques, we assume that singular values are similar, i.\,e. $\b\Sigma_s^l \simeq \b\Sigma_t^l$ and fix them. Naturally, this assumption does not always hold. See Sec.~\ref{sec:sampling} for a discussion.
\textit{If they are fixed, then} the optimal solution to~\eqref{eq:transfer_opt_problem_subspace} is easily obtained by solving the linear equation and obtain the solution $\*M = \*U_t^l{\*U_s^l}^T$. By applying $\*M$ to~\eqref{eq:transfer_opt_problem_subspace} the source data becomes
\begin{equation}\label{eq:non_approximated_source_solution}
    \*X_s^l=\*M\*U_s^l \b\Sigma_s^l = \*U_t^l{\*U_s^l}^T\*U_s^l \b\Sigma_s^l = \*U_t^l\b\Sigma_s^l \in \mathbb{R}^{n\times l}
\end{equation}
and is used for training an invariant classifier. The resulting model can be evaluated on $\*X_t^l=\*U_t^l\b\Sigma_t^l \in \mathbb{R}^{n\times l}$. This overrides the source basis and prevents the source subspace to be arbitrarily different from the target due to the affiliation to the target space. The solution also fulfills the constrains because $\*M$ is an orthogonal matrix due to the orthogonal matrices $\*U_t^l$ and ${\*U_s^l}^T$. In particular,~\eqref{eq:non_approximated_source_solution} projects the source data onto the principal components of the subspace basis of $\*X_t$.
If data matrices $\*X_t$ and $\*X_s$ are standardized, the geometric interpretation is a rotation of source data w.r.t to angles of the target basis. We call this procedure Subspace Override (SO).

This procedure requires a complete eigenspectrum and scales to $\mathcal{O}(n^3)$ in worst case~\cite{NIPS2000_1866}. Further, all available data is required for this approach. Using Nyström techniques, we show that only a subset of the data is required, which simultaneously reduces computational complexity and eliminates the need to examine all singular values.

\subsection{Nyström Extension}\label{sec:NSO}
For clarity, the following notation will overlap with the previous section but keeps things simple. We assume the reader is familiar with Nyström SVD techniques. Otherwise, the reader may consider Appendix \ref{sec:preliminaries} for an introduction to the Nyström approximation.

In short, the Nyström SVD technique is a low-rank approximation which decomposes a given matrix $\mathbf{K} \in \mathbb{R}^{n\times d}$ into the constitution
\begin{equation}\label{eq:matrix_decomposition_short}
    \mathbf{K}=
    \begin{bmatrix}
        \mathbf{A} & \mathbf{B} \\
        \mathbf{C} & \mathbf{F} \\
    \end{bmatrix},
\end{equation}
with $\*{A} \in \mathbb{R}^{l\times l}$, $\*{B} \in \mathbb{R}^{l\times (d-l)}$, $\*{C} \in \mathbb{R}^{(n-l)\times l} $ and $\*{F} \in \mathbb{R}^{(n-l)\times (d-l)}$. The matrix $\*A$ contains the random samples called the landmark matrix. Given $\*K$, the singular value decomposition $\*A = \*U \b\Sigma \*V^T$, and $\*C$, the full SVD of $\*K$ is reconstructable, which is similar to the following approach.

Consider $\*X_s$ and $\*X_t$ with the decomposition as in~\eqref{eq:matrix_decomposition_short}.
For a Nyström SVD, we sample from both matrices $l$ rows/columns obtaining landmarks matrices $\*A_s = \*U_s \b\Sigma_s{\*V_s}^T  \in \mathbb{R}^{l\times l}$ and $\*A_t = \*U_t \b\Sigma_t {\*V_t}^T \in \mathbb{R}^{l\times l}$.
The target data is projected into the subspace as in~\eqref{eq:proof_valid_pca} via the Nyström technique (Appendix~\ref{sec:preliminaries}) and keeps only the most relevant data structures via
\begin{equation}\label{eq:ny_test_approx}
    \tilde{\*X}_t^l = \tilde{\*U}_t \b\Sigma_t =
    \begin{bmatrix}
        \*U_t \\
        \hat{\*U}_t
    \end{bmatrix}
    \b\Sigma_t
    =
    \begin{bmatrix}
        \*U_t \\
        \*C_t \*V_t \b\Sigma_t^{-1}
    \end{bmatrix} \b\Sigma_t \in \mathbb{R}^{n\times l}.
\end{equation}
Analogously, the source data could be approximated by  $\*X_s^l =  \tilde{\*U}_s \b\Sigma_s  \in \mathbb{R}^{n\times l}$.
The Nyström technique is also used to approximate the solution to the optimization problem with $\*M = \tilde{\*U}_t \tilde{\*U}_s^{-1}$ and project the source data into the target subspace via
\begin{equation}\label{eq:ny_training_approx}
    \tilde{\*X}_s^l = \*M \tilde{\*U}_s \b\Sigma_s = \tilde{\*U}_t \tilde{\*U}_s^{-1}\tilde{\*U}_s \b\Sigma_s =\tilde{\*U}_t \b\Sigma_s  \in \mathbb{R}^{n\times l}.
\end{equation}
Hence, it is sufficient to only compute a \ac{SVD} of $\*A_t$ and $\*A_s$ instead of $\*X_t$ and $\*X_s$ with $l\ll m,d,n$ and therefore is considerably lower in computational complexity.

By definition of the Nyström approximation, it is $\tilde{\*U}_s\tilde{\*U}_s^T = \tilde{\*U}_t\tilde{\*U}_t^{-1} =\*I$ and $\tilde{\*U}_t$ is an orthogonal basis.
Therefore, the subspace projections are orthogonal transformations and fulfill the constrains of~\eqref{eq:transfer_opt_problem_subspace}.

{Besides small sample requirements, the major advantage of using the approximated low-rank solution in favor of the optimal solution is that singular values that are closer to zero are set to zero, reducing the noise of the data in the subspace. Therefore the approach focuses on intrinsic data characteristics, which should lead to better classification performance.}

Subsequently, this approach is denoted as Nyström Subspace Override (NSO).
The matrix $\*X_s^l$ is used for training, and $\*X_t^l$ is used for testing.
But uniform sampling may not be optimal for Nyström, given a classification task~\cite{DBLP:journals/ijon/SchleifGT18}. Therefore, we subsequently integrate class-wise sampling in the following. Pseudo code shown in Algorithm~\ref{alg:pseudocodentm}.
\subsection{Sampling Strategy}\label{sec:sampling}
The standard technique to create Nyström landmark matrices is to sample uniformly or find clusters in the data~\cite{Kumar.2012}. In supervised classification with more than two classes, class-wise sampling should be utilized to properly include class-depending attributes of a matrix into the approximation~\cite{DBLP:journals/ijon/SchleifGT18}.
However, a decomposition as in~\eqref{eq:matrix_decomposition}, required for Nyström SVD, is intractable with class-wise sampling, because respective matrices are non-square.
Let $\*X_s \in \mathbb{R}^{m\times d}$ with $m \neq d$ and landmark indices $I=\{i_1,\dots,i_s\}$ with at least one $i_j>d$ and if $m>d $, then it is undefined. Therefore, we sample rows class-wise and obtain $\*A_s^d \in \mathbb{R}^{l\times d}$ instead of $\*A_s \in \mathbb{R}^{l\times l}$, making it possible to sample from the whole range of source data.
The sampling from test data $\*X_t$ is done uniformly row-wise, because of missing class information.
The resulting singular value decompositions, i.\,e. $\*A_t^d =\*U_t^d \b\Sigma_t^d{\*V_t^d}^T$ and $\*A_s^d =\*U_s^d \b\Sigma_s^d{\*V_s^d}^T$, are utilized for successive Nyström approximations.

However, the possible numerical range of $\b\Sigma_{(\cdot)}^d$ and $\b\Sigma_{(\cdot)}$ is naturally not the same, which is easily shown by the Gerschgorin Bound (Theorem~\ref{theo:discs} in Appendix~\ref{app:gerschgorin}). It scales approximated matrices $\*X_{(\cdot)}^l$ different by $\b\Sigma_{(\cdot)}^d$ and accurate scaling of the singular vectors cannot be guaranteed. Therefore, we apply a post-processing correction and standardize the approximated matrices to transform the data back to mean zero and variance one.
The singular vectors also have an approximation error. However, both subspace projections are based on the same transformation matrix, hence making an identical error, and as a result, the error should not affect the classification.

\begin{figure}
    \centering
\begin{subfigure}[b]{.32\textwidth}
    \includegraphics[width=1\textwidth]{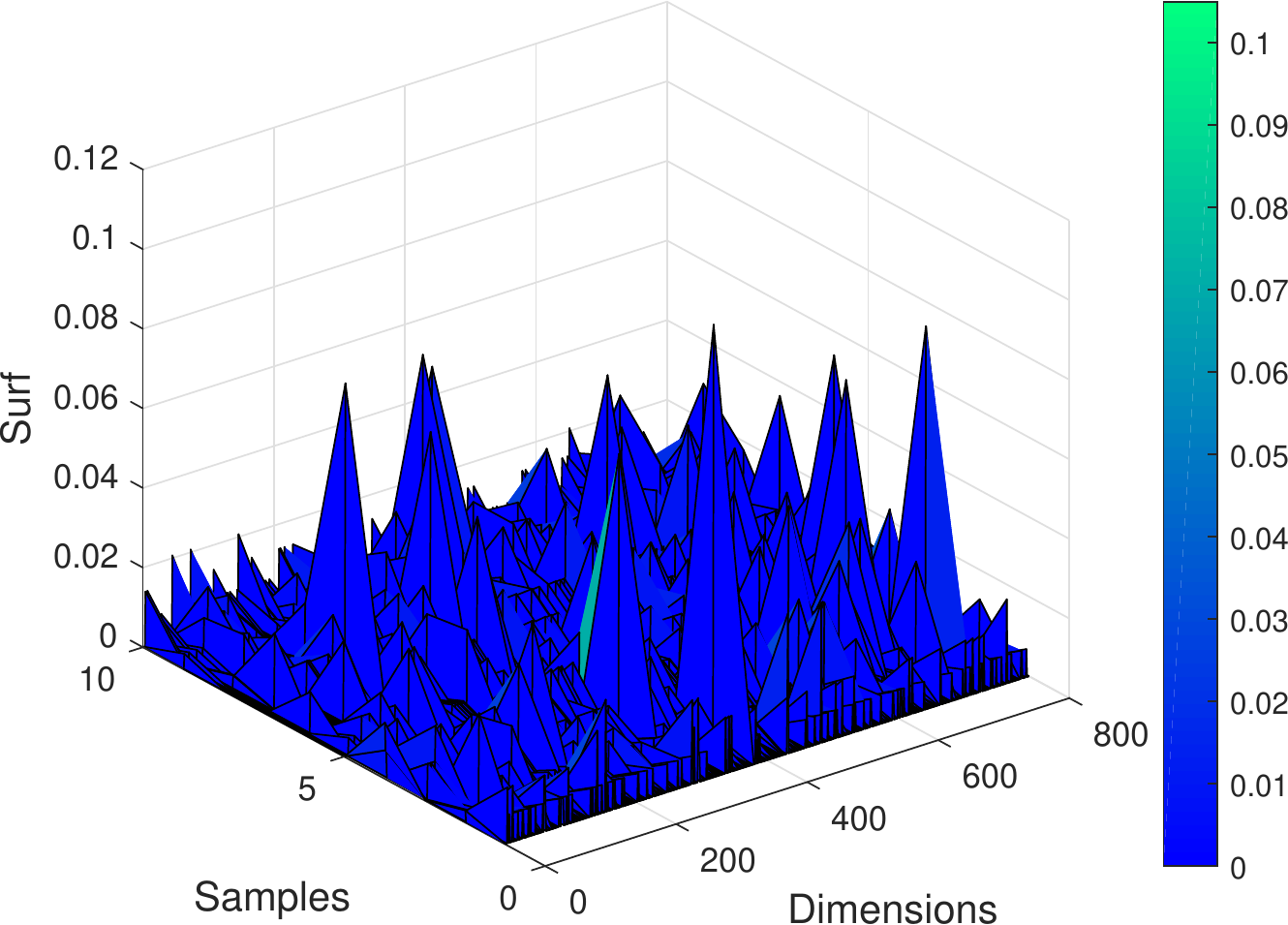}
    \subcaption{Target Samples}
    \label{Fig:TargetSample}
\end{subfigure}
\begin{subfigure}[b]{.32\linewidth}
    \includegraphics[width=1\textwidth]{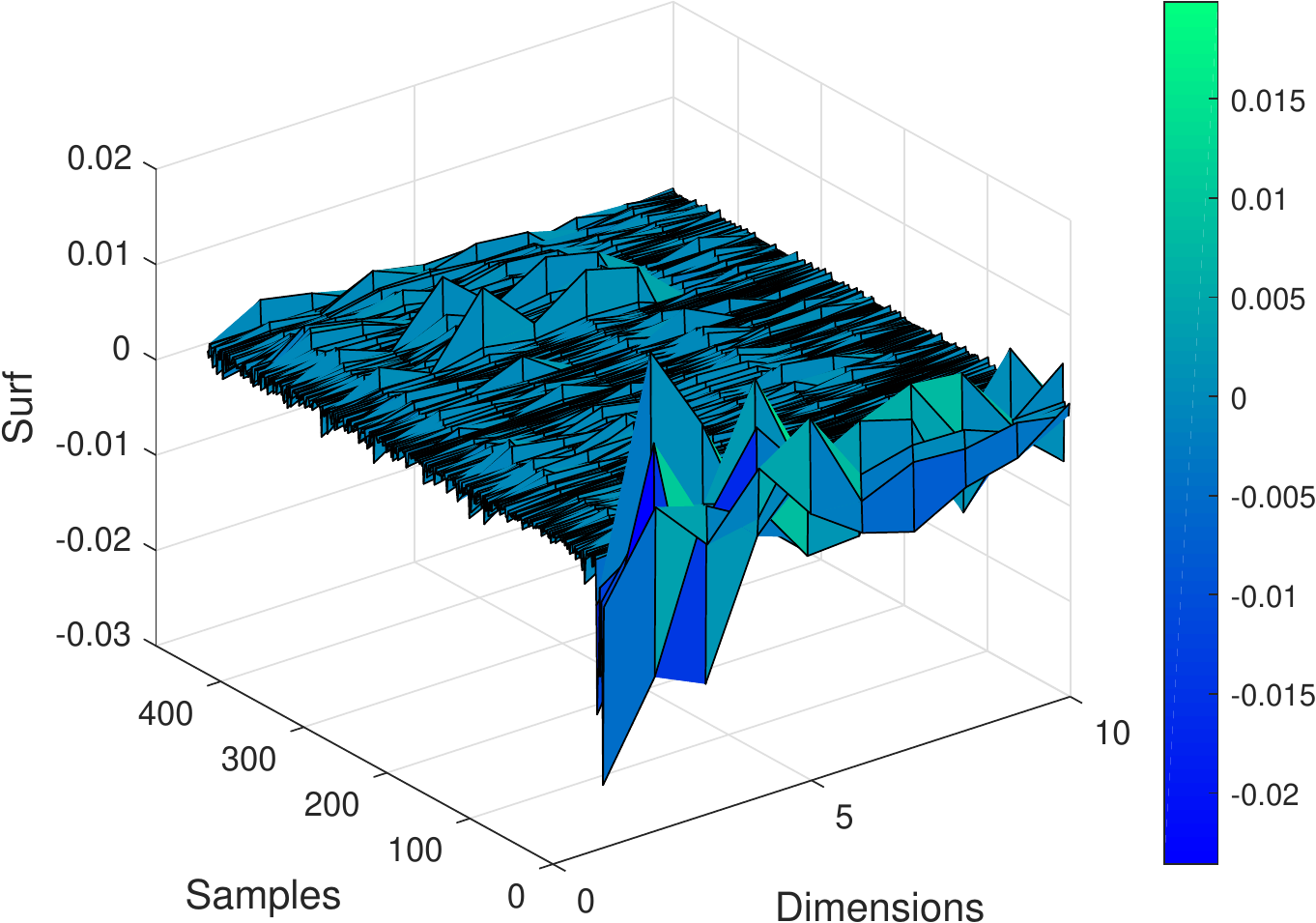}
    \subcaption{Target in Subspace}
    \label{Fig:TargetNSO}
\end{subfigure}
\begin{subfigure}[b]{.32\linewidth}
    \includegraphics[width=1\textwidth]{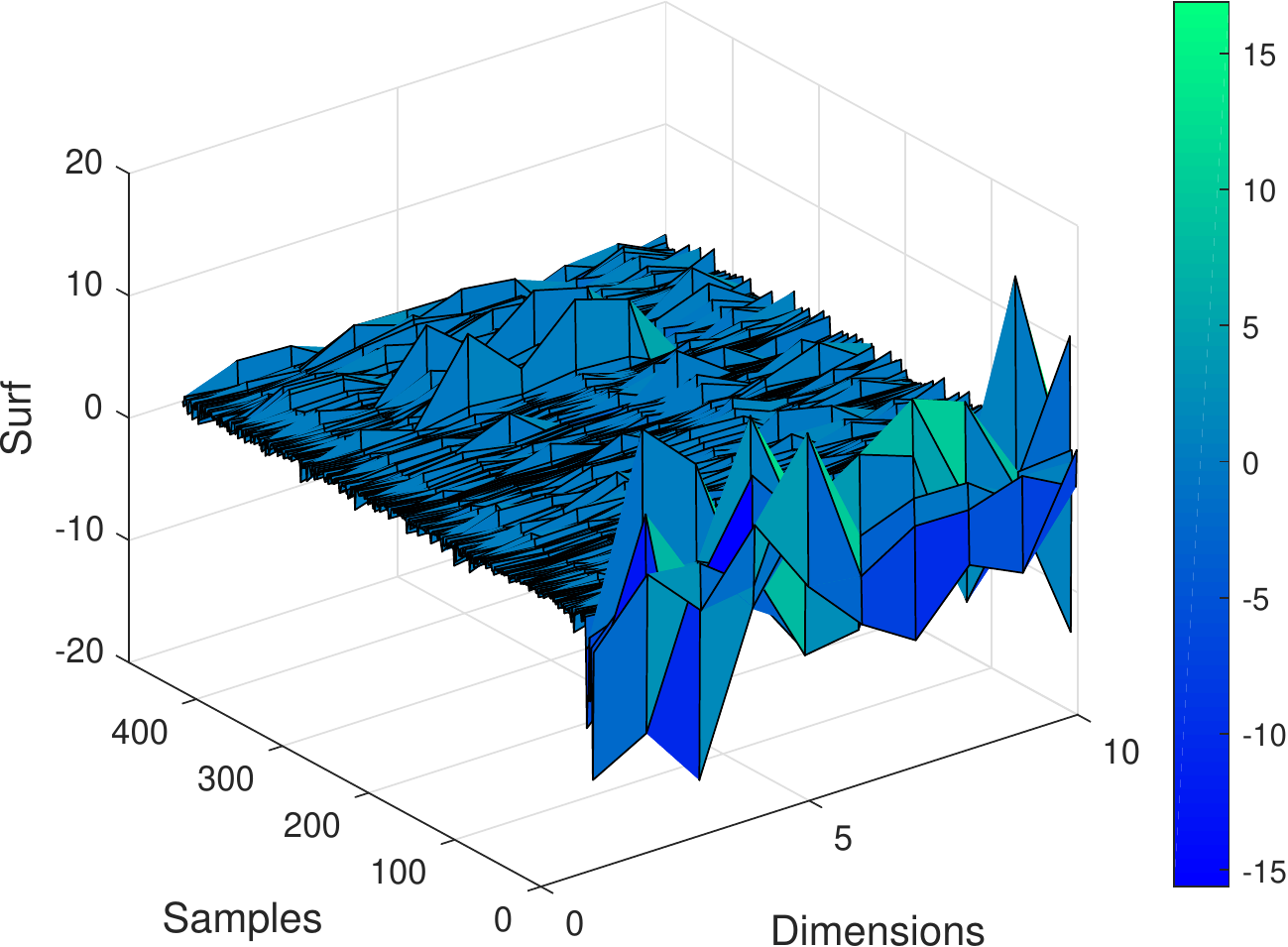}
    \subcaption{Standardization of Target Subspace}
    \label{Fig:TargetNorm}
\end{subfigure}
\\
\begin{subfigure}[b]{.32\linewidth}
    \includegraphics[width=1\textwidth]{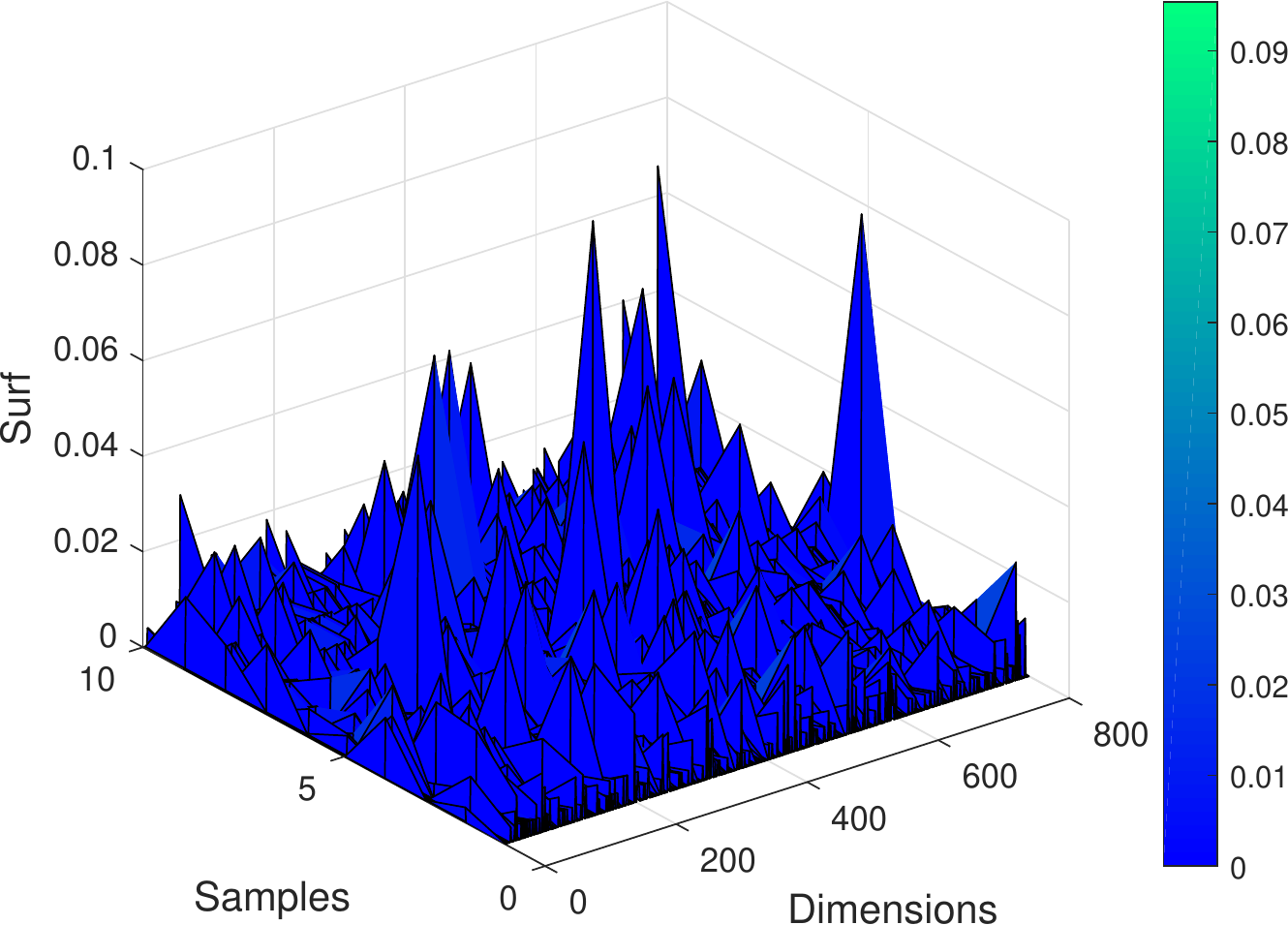}
    \subcaption{Source Samples}
    \label{Fig:SourceSample}
\end{subfigure}
\begin{subfigure}[b]{.32\linewidth}
    \includegraphics[width=1\textwidth]{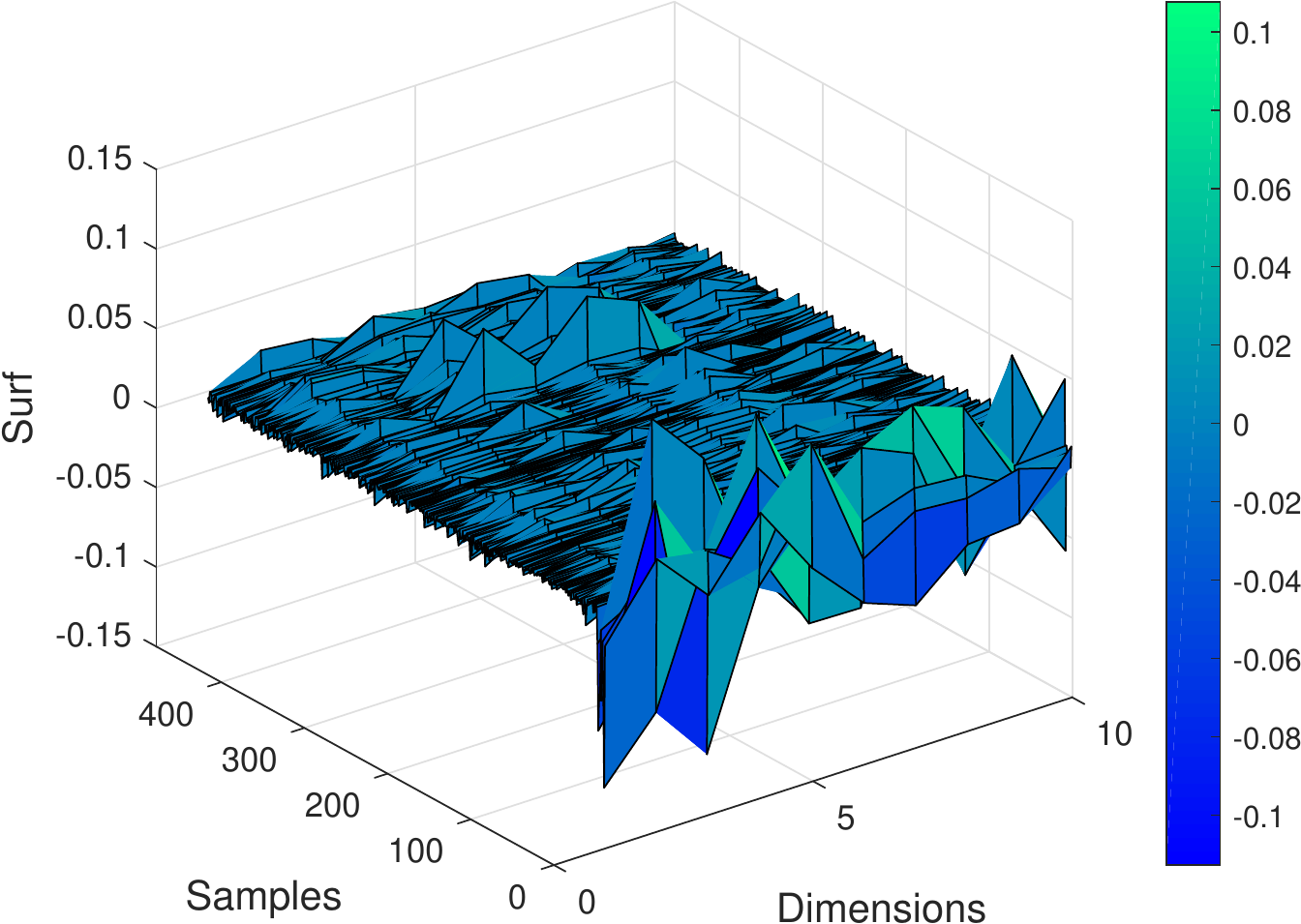}
    \subcaption{Source in Subspace}
    \label{Fig:SourceNSO}
\end{subfigure}
\begin{subfigure}[b]{.32\linewidth}
    \includegraphics[width=1\textwidth]{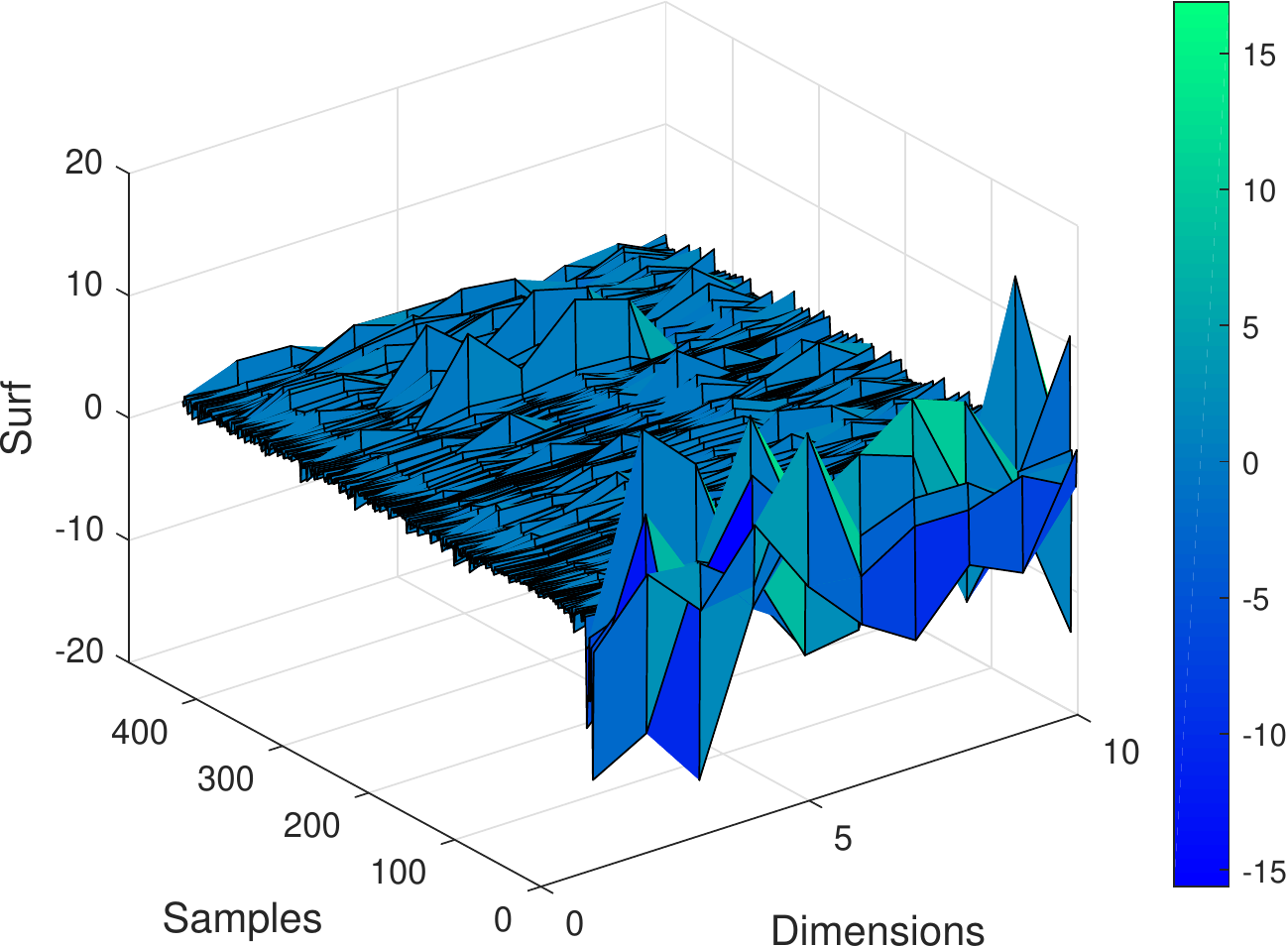}
    \subcaption{Standardization of Source Subspace}
    \label{Fig:SourceNorm}
\end{subfigure}
\caption{Process of \textit{Nyström Subspace Override} with ten landmark samples applied to \textit{Caltech vs Amazon} image dataset encoded with surf features as a surface plot. Best viewed in color.\label{FigBtProcess}}
\end{figure}

The process of Nyström Subspace Override (NSO) is given in Fig.~\ref{FigBtProcess}. The first column visualizes the samples of Nyström to create the approximated set of subspace projection operators. The second column shows the data after the subspace projection. The similarity in structure but dissimilarity in scaling, as discussed above, is visible. The last column shows the data after applying post-correction and leading to a high similarity afterward.
The pseudo code of NSO is shown in Algorithm~\ref{alg:pseudocodentm}.
\begin{algorithm}
    \caption{Nyström Subspace Override}\label{alg:pseudocodentm}
    \begin{algorithmic}[1]
        \Require $\*X_s$ as $m$ sized training; $\*X_t$ as $n$ sized test set; $\mathbf{Y}$ as $m$ sized training label vector; $l$ as number of landmarks parameter.
        \Ensure New Source $\tilde{\*X}_s^l$; new Target $\tilde{\*X}_t^l$;
        \State $ \*X_s,\*Y_s$ = augmentation($\*X_s$,$\*Y_s$,$n$)  \\  \Comment{Gaussian sampling or random removal to make $\*X_s$ equally sized to $\*X_t$.}
        \State $\*A_t^d,\*A_s^d,\*C_t$ = decomposition($\*X_t$,$\*X_s$,$\*Y$,$l$)   \Comment{Eq.\eqref{eq:matrix_decomposition_short}}
        \State  $ \b\Sigma_s^d$ = $SVD(\mathbf{A}_s^d)$; 
        \State  $ \*U_t^d, \b\Sigma_t^d, {\*V_t^d}$ = $SVD(\mathbf{A}_t^d)$; 
        \State   $\tilde{\*U}_t$ =  $\begin{bmatrix}
        \*U_t^d &
        \*C_t {\*V_t^d} {\b\Sigma_t^d}^{-1}
    \end{bmatrix}^T$ \Comment{Eq.~\eqref{eq:ny_test_approx}}
        \State $\tilde{\*X}_t^l  = \tilde{\*U}_t \b\Sigma_t^d $ \Comment{Eq.~\eqref{eq:ny_test_approx}}
        \State $\tilde{\*X}_s^l  = \tilde{\*U}_t \b\Sigma_s^d $ \Comment{Eq.~\eqref{eq:ny_training_approx}}
        \State $ \tilde{\*X}_s^l,\tilde{\*X}_t^l $=standardization($\tilde{\*X}_s^l$,$\tilde{\*X}_t^l$) \Comment{Effect as in Fig.~\ref{FigBtProcess}}
    \end{algorithmic}
\end{algorithm}
\subsection{Properties of Nyström Subspace Override}\label{sec:properties}
The computational complexity of \ac{NSO} is composed of economy-size \ac{SVD} of landmark matrices $\*A_s^d$ and $\*A_t^d$ with complexity $\mathcal{O}(2l^2)$.
The matrix inversion of diagonal matrix ${\b\Sigma_t^d}^{-1}$ in~\eqref{eq:ny_test_approx} can be neglected.
The remaining $k$ matrix multiplications are of complexity $\mathcal{O}(kl^2)$ and are therefore contributing to the overall complexity of \ac{NSO}, which is $O(l^2)$ with $l\ll n,m,d$. This makes \ac{NSO} the fastest subspace domain adaptation solution in terms of computational complexity in comparison to \textit{compared methods in Sec.~\ref{sec:experiments}}.

The \textbf{out-of-sample extension} for unseen target/source samples, e.\,g. $\*x \in \mathbb{R}^d$, is analog to~\eqref{eq:ny_test_approx}. Based on~\eqref{eq:proof_valid_pca}, a subspace projection via (approximated) right singular vectors is also valid. Hence, a sample can be projected into the subspace via  \begin{equation}
    \*x^l = \*x \tilde{\*V}_t^T = x
    \begin{bmatrix}
        \*V_t  && \b\Sigma_t^{-1} \*U^T \*B_t
    \end{bmatrix}
\end{equation}
and be evaluated by an arbitrary classifier learned in the subspace.

The difference between source and target domain after SO, i.\,e. approximation error of source by target domain is bounded by
\begin{equation}\label{eq:error}
    E_{SO} = \norm{\*X_s^l-\*X_t^l} < \sum_{i=1}^{l+1} (\sigma_i(\*X_s) - \sigma_i(\*X_t))^2 < \norm{\*X_s-\*X_t}.
\end{equation}
Where $\sigma_i(\cdot)$ is the $i$-th singular value in descending order of $\*X_s$ and $\*X_t$ respectively and $1<l< min(n,d)$. The proof can be found in Appendix~\ref{sec:proof}.
As in prior LS approaches~\cite{Fernando2013a,sun2016return,Aljundi2015}, we want NSO to minimize the difference between the source and target data. In Eq.~\eqref{eq:error} is shown that NSO has a lower norm to the original data and proves that the matrices are aligned during NSO, making them numerically more similar. Note that similar matrices not necessarily indicate a good classification performance in terms of accuracy by an arbitrary classifier in a domain adaptation setting. The classification performance is evaluated in the following.
\section{Experiments}\label{sec:experiments}
We follow the experimental design typical for domain adaptation algorithms~\cite{Dai2007,Long2015,Gong2012,Long2013a,Long2014a,5288526,Pan2011,sun2016return,Mahadevan19,Fernando2013a,Aljundi2015,8100030}. The tests are conducted on the common datasets Reuters, Newsgroup and Office-Caltech. A crucial characteristic of datasets for domain adaptation is that domains for training and testing are different but related, e.\,g. sharing the same categories. The NSO approach is evaluated against the common and state of the art domain adaptation methods TCA~\cite{Pan2011}, GFK~\cite{Gong2012}, JDA~\cite{Long2013a}, SA~\cite{Fernando2013a}, CORAL~\cite{sun2016return}, EasyTL\cite{WangEasyTL}, SCA~\cite{Ghifary2017}, MEDA~\cite{meda2018} and JGSA~\cite{8100030}. We extend the object detection study by also evaluating against deep DA networks. We follow~\cite{meda2018} and use the Alexnet~\cite{Krizhevsky2012} as the baseline for Deep-Coral\cite{Baochen2016}, JAN\cite{LongJAN2017}, DAN~\cite{Long2018} and DDC\cite{Tzeng2014}. The networks are always trained on original images.
The parameters for the respective method are determined for the best performance in terms of accuracy via grid search.
In the experiments, the \ac{SVM} independent of being a baseline or underlying classifier for domain adaptation methods uses the RBF-Kernel. All experiments are done via the standard sampling protocol~\cite{LongJAN2017} and use all available source and target data. We did 20 test runs and summarized the result as mean accuracy.
\subsection{Dataset Description}\label{sec:description}
A summary of all datasets is shown in Tab.~\ref{TableSumData}.
Regardless of the dataset, it has been standardized to standard mean and variance.\newline
\textbf{Reuters-21578}~\cite{Dai2007}: A collection of Reuters news-wire articles collected in 1987 as TFIDF features.
The three top categories \textit{organization (orgs)}, \textit{places} and \textit{people} are used in our experiment.\\
To create a transfer problem, a classifier is not tested with the same categories as it is trained on, e.\,g. it is trained on some subcategories of organization and people and tested on others.
Six datasets are generated: \textit{orgs vs. places}, \textit{orgs vs. people}, \textit{people vs. places}, \textit{places vs. orgs}, \textit{people vs. places} and \textit{places vs. people}.
They are two-class problems with the top categories as the positive and negative class and with subcategories as training and testing examples.

\textbf{20-Newsgroup}~\cite{Long2014a}: The original collection has approximately 20.000 text documents from 20 Newsgroups and is nearly equally distributed in 20 subcategories.
The top four categories are \textit{comp}, \textit{rec}, \textit{talk} and \textit{sci}, each containing four subcategories.
We follow a data sampling scheme introduced by~\cite{Long2015} and generate 216 cross domain datasets based on subcategories, which are summarized as mean over all test runs as \textit{comp vs rec}, \textit{comp vs talk}, \textit{comp vs sci}, \textit{rec vs sci}, \textit{rec vs talk} and \textit{sci vs talk}.

\textbf{Caltech-Office (OC)}~\cite{Gong2012}:
The first, Caltech (\textit{C}), is an extensive dataset of images and contains 30.607 images within 257 categories.
The Office dataset is a collection of images drawn from three sources, which are from \textit{amazon (A)}, digital SLR camera \textit{(DSLR)} and \textit{webcam (W)}. They vary regarding camera, light situation and size, but ten similar object classes, e.\,g. computer or printer, are extracted for a classification task. We use SURF~\cite{Gong2012} and DeCaf\cite{Donahue2014} features.
\begin{table}
    \caption{Overview of the dataset characteristics containing numbers of samples, features and labels. }
    \label{TableSumData}
    \centering
    \resizebox{\textwidth}{!}{%
    \begin{tabular}{c|c|c|c|c}
            \textbf{Dataset}    & \textbf{Subsets}       & \textbf{\#Samples}  & \textbf{\#Features}      & \textbf{\#Classes}     \\ \hline
            Caltech             & C                      & 1123                &  800 (4096)             &  10          \\
            Office              & A,W,D                  & 1123                &  800 (4096)             &  10          \\
            Newsgroup           & Comp,Rec,Sci,Talk      & 4857,3967,3946,3250 &  25804                  &  2           \\
            Reuters             & Orgs,People,Places     & 1237,1208,1016      &  25804                  &  2           \\
    \end{tabular}}
\end{table}
\subsection{Performance Results}\label{sec:results_performance}
\begin{figure}[b]
    \centering
    \begin{subfigure}[c]{0.3\columnwidth}
        \includegraphics[width=1\columnwidth]{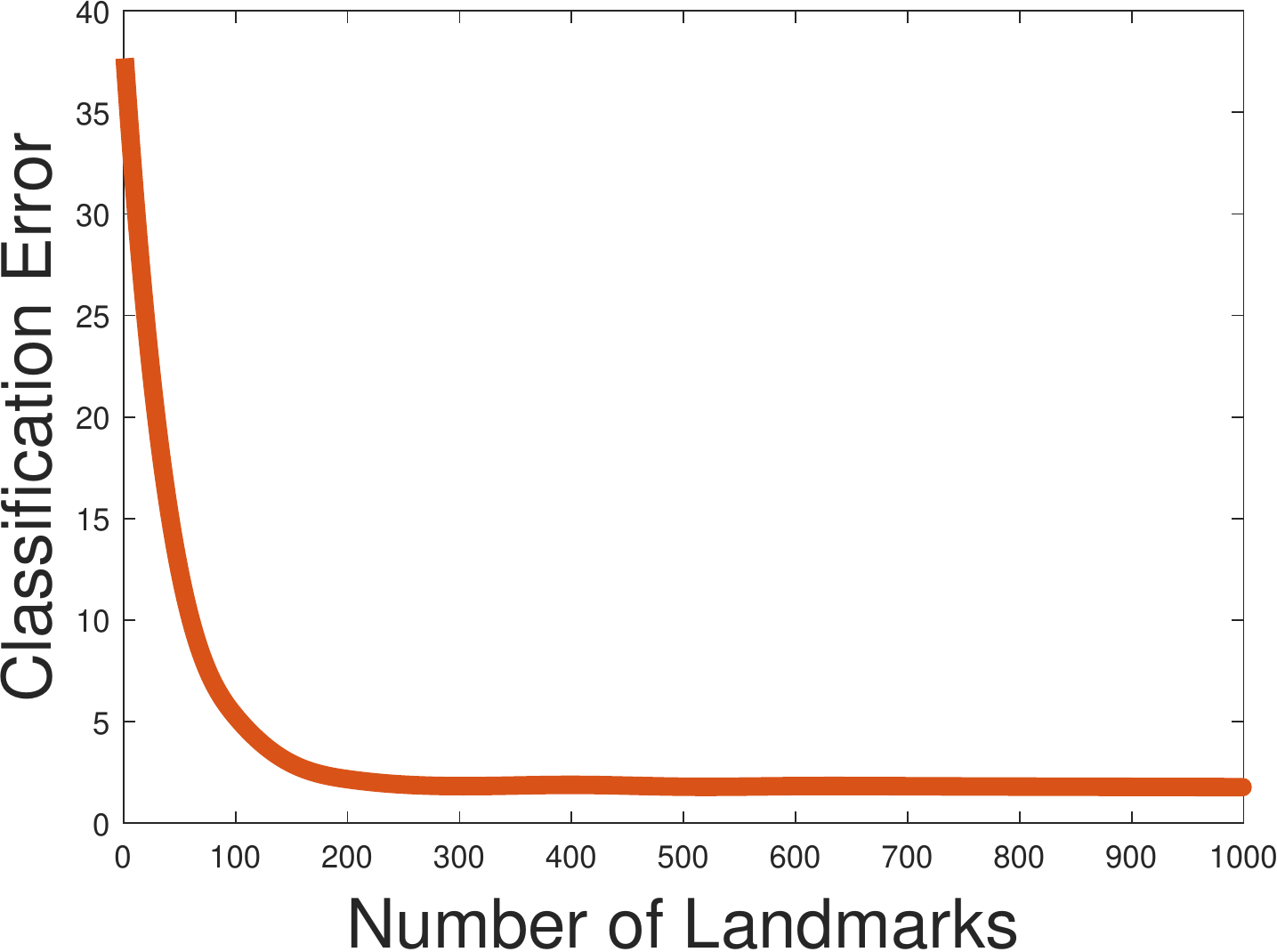}
        \subcaption{Reuters\label{FigTextEL}}
    \end{subfigure}
    \begin{subfigure}[c]{0.3\columnwidth}
     \includegraphics[width=1\columnwidth]{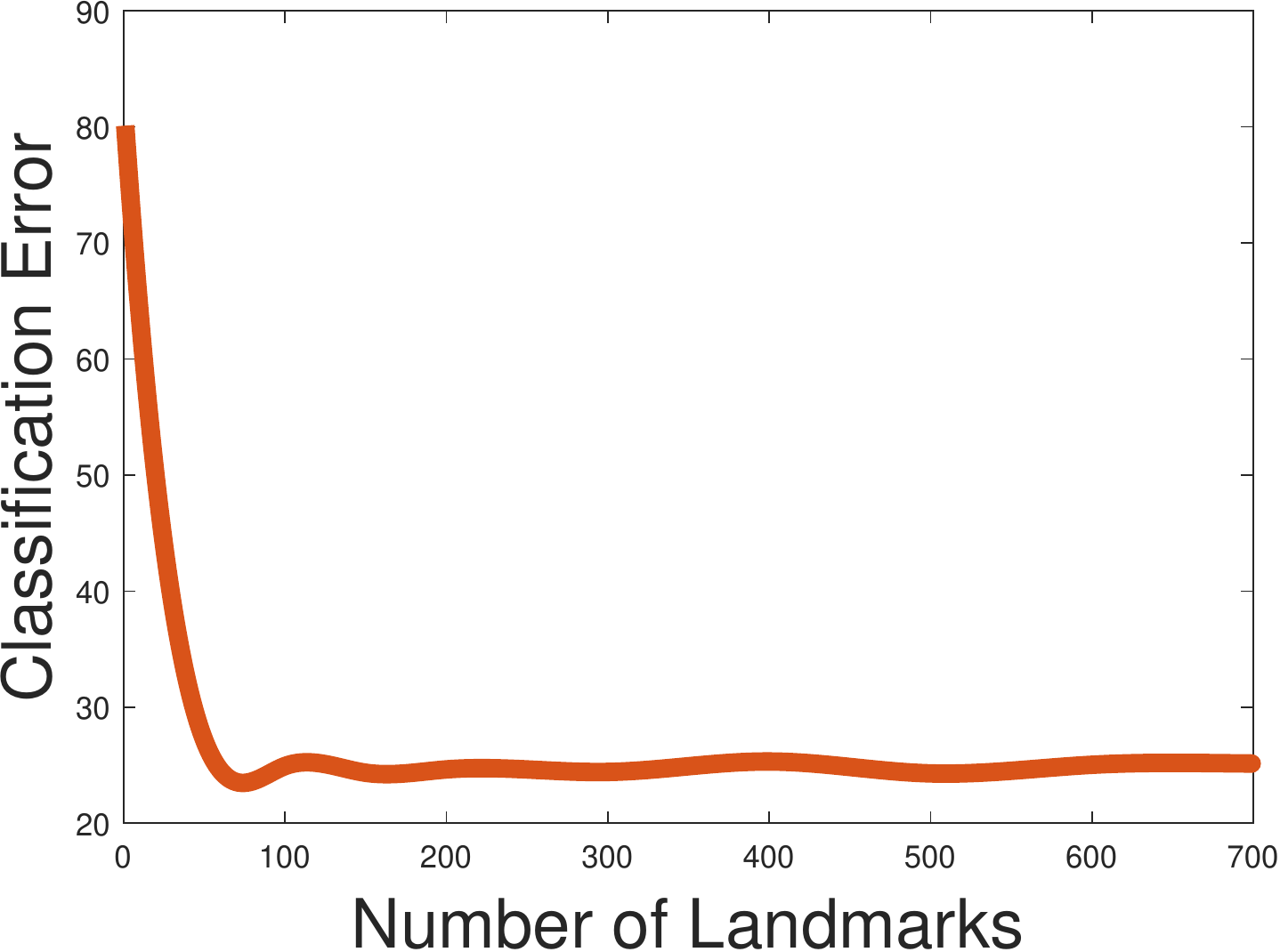}
        \subcaption{OC - Surf\label{FigImageEL}}
    \end{subfigure}
    \begin{subfigure}[c]{0.3\columnwidth}
        \includegraphics[width=1\columnwidth]{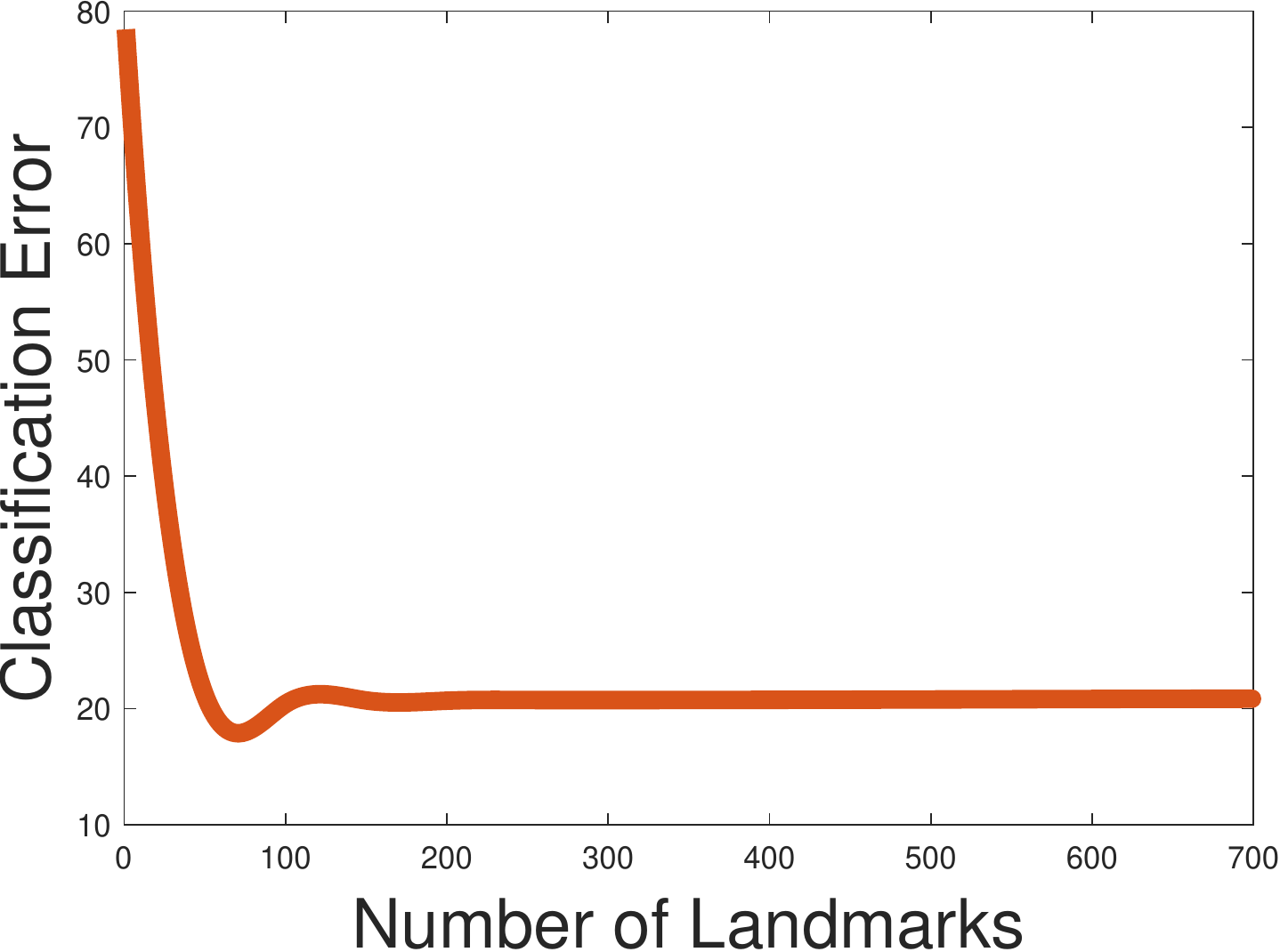}
           \subcaption{OC - DeCaf \label{FigImageDecafEL}}
       \end{subfigure}
    \caption{Relationship between the number of landmarks and mean error on Reuters and  Office-Caltech datasets.\label{FigErrorvslandmarks}}
\end{figure}

The results are shown per dataset separately. The results on Newsgroup in Tab.~\ref{tab:result_newsgroup}, Reuters in Tab.~\ref{tab:result_reuters}, OC with Surf features in Tab.~\ref{tab:result_caltech_surf}, OC with decaf and deep DA methods in Tab.~\ref{tab:result_caltech_decaf}. Summarizing, our NSO algorithm is basically the best on Reuters and Newsgroup data. The only competitive algorithm is SA on Reuters data with similar results to ours. SA is also an LS subspace approach. However, SA is outperformed by NSO at Newsgroup. NSO demonstrates its usefulness for large sparse matrices that are given at these datasets. At the OC-Surf dataset, the NSO outperforms on many datasets and has the best mean accuracy. Only at OC-Decaf features, NSO is midfield in performance, but it is still competitive. We assume that the Decaf features are very dense feature matrices in terms of descriptive information even if the singular values are small. Therefore, the low-rank approximation is contra-productive.

The intriguing part of this evaluation comes with the cross-task evaluation. While SA is very good at Reuters and Newsgroup, it has bad performance on OC datasets. While MEDA and JGSA have poor performance at Reuters and Newsgroup, they are good at OC datasets. Our NSO approach is in three out of four tasks the recommendable choice showing convincing task-independent performance. In Fig.~\ref{FigErrorvslandmarks}, the parameter sensitivity is shown and demonstrates that the parameterization (number of landmarks) of NSO is stable, simple to optimize and supports the Nyström error expectation.
\begin{table}
    \caption{Mean accuracy of traditional DA methods on Newsgroup text dataset.}\label{tab:result_newsgroup}
    \resizebox{\textwidth}{!}{%
\begin{tabular}{l|cccccccccccc}
    \textbf{Dataset}    &    SVM &   TCA  &   JDA  &   GFK &  SA    & CORAL  &  CGCA  &    SCA & EasyTL & JGSA &   MEDA & NSO (ours)\\ \hline
 Comp vs Rec            & 77.6  &  78.9  &  83.1  &  75.1  &  78.5  &  79.4  &  84.0  &  56.1  &  42.2  &  88.3  &  49.1  &  \textbf{90.2}\\
 Comp vs Sci            & 71.1  &  62.0  &  75.5  &  64.1  &  80.2  &  71.8  &  73.2  &  72.4  &  25.2  &  78.4  &  49.2  &  \textbf{98.4}\\
 Comp vs Talk           & 84.4  &  75.0  &  87.7  &  83.8  &  91.1  &  90.5  &  87.0  &  89.5  &  41.1  &  91.2  &  54.4  &  \textbf{96.7}\\
 Rec vs Sci             & 69.3  &  79.6  &  79.0  &  64.4  &  81.1  &  75.0  &  74.0  &  71.4  &  33.8  &  80.5  &  50.0  &  \textbf{99.0}\\
 Rec vs Talk            & 74.5  &  86.6  &  82.0  &  72.9  &  79.7  &  81.6  &  77.3  &  78.3  &  41.6  &  80.8  &  55.0  &  \textbf{96.4}\\
 Sci vs Talk            & 70.9  &  77.6  &  70.5  &  64.2  &  76.0  &  74.2  &  69.0  &  72.2  &  41.1  &  77.7  &  53.7  &  \textbf{96.4}\\ \hline
  Mean                  & 74.6  &  76.6  &  79.6  &  70.7  &  81.1  &  78.7  &  77.4  &  73.3  &  37.5  &  82.8  &  51.9  &  \textbf{96.2}\\
\end{tabular}}
\end{table}
\begin{table}
    \caption{Mean accuracy of traditional DA methods on Reuters text dataset.\label{tab:result_reuters}}    \resizebox{\textwidth}{!}{%
\begin{tabular}{l|cccccccccccc}
    \textbf{Dataset}     &    SVM &   TCA &   JDA &   GFK &    SA & CORAL &  CGCA &    SCA & EasyTL & JGSA &   MEDA & NSO (ours)\\ \hline
 Orgs vs People  &   78.1 &  79.5 &  76.6 &  75.3 &  \textbf{99.9} &  77.5 &  78.0 &  77.8  &   39.2 & 76.5 &  48.0  & 99.6 \\
 People vs Orgs  &   79.2 &  82.7 &  80.0 &  71.6 &  \textbf{99.9} &  78.2 &  78.6 &  79.8  &   37.9 & 74.2 &  47.3  & 98.5 \\
 Orgs vs Place   &   69.2 &  72.9 &  70.0 &  60.5 &  97.3 &  70.3 &  70.1 &  69.8  &   28.9 & 72.2 &  43.2  & \textbf{98.6 }\\
 Place vs Orgs   &   66.3 &  71.1 &  65.6 &  61.5 &  \textbf{97.2} &  66.5 &  67.7 &  65.3  &   27.0 & 64.4 &  41.4  & \textbf{97.2} \\
 People vs Place &   55.7 &  57.4 &  57.0 &  57.5 &  \textbf{97.4} &  57.8 &  57.0 &  57.3  &   22.4 & 52.6 &  40.9  & \textbf{97.4} \\
 Place vs People &   57.4 &  48.9 &  60.7 &  56.2 &  \textbf{97.4} &  56.3 &  54.4 &  58.2  &   18.3 & 55.5 &  38.5  & \textbf{97.4 }\\\hline
  Mean          &    67.7 &  68.7 &  68.3 &  63.8 &  \textbf{98.1} &  67.7 &  67.6 &  68.0  &   28.9 & 65.9 &   43.2 & \textbf{98.1} \\
\end{tabular}}
\end{table}
\begin{table}
    \centering
    \caption{Mean accuracy of traditional DA on Caltech-Office with surf features.\label{tab:result_caltech_surf}}
    \resizebox{\textwidth}{!}{%
    \begin{tabular}{l|cccccccccccc}
        \textbf{Dataset}     & SVM   & TCA   & JDA   &  GFK  & SA  & CORAL & CGCA & SCA &EasyTL & JGSA &MEDA & NSO (ours)\\ \hline
        C vs A      &  53.1 &  53.9 &  55.2 &  41.8 &52.2 &  52.1 &  54.1 &  33.1 &  50.1 &  51.8 &  56.5 & \textbf{88.5} \\
        C vs W      &  41.7 &  42.4 &  46.8 &  40.7 &18.3 &  38.6 &  43.1 &  24.9 &  49.5 &  46.1 &  53.9 & \textbf{81.0} \\
        C vs D      &  47.8 &  46.5 &  49.7 &  39.5 &15.9 &  36.3 &  37.6 &  33.1 &  48.4 &  44.6 &  50.3 & \textbf{79.0} \\
        A vs C      &  41.7 &  45.4 &  43.5 &  39.0 &60.0 &  45.1 &  44.9 &  26.3 &  43.0 &  39.7 &  43.9 & \textbf{61.5} \\
        A vs W      &  31.9 &  37.6 &  44.4 &  36.9 &29.2 &  44.4 &  43.9 &  27.6 &  40.7 &  46.1 &  53.2 & \textbf{81.0} \\
        A vs D      &  44.6 &  40.1 &  31.2 &  33.1 &28.0 &  39.5 &  36.3 &  25.5 &  38.9 &  47.8 &  45.9 & \textbf{79.0} \\
        W vs C      &  21.2 &  31.2 &  31.5 &  27.4 &23.2 &  33.7 &  33.8 &  15.6 &  29.7 &  30.2 &  34.2 & \textbf{63.5} \\
        W vs A      &  27.6 &  34.7 &  31.7 &  31.2 &29.5 &  35.9 &  37.6 &  21.1 &  35.2 &  40.0 &  42.7 & \textbf{95.8} \\
        W vs D      &  78.3 &  83.4 &  92.4 &  82.8 &78.3 &  86.6 &  88.5 &  41.4 &  77.1 &  \textbf{91.1} &  88.5 & 79.0 \\
        D vs C      &  26.5 &  36.2 &  32.6 &  27.2 &21.9 &  33.9 &  35.4 &  17.2 &  31.3 &  30.3 &  34.8 & \textbf{66.6} \\
        D vs A      &  26.2 &  37.1 &  36.7 &  30.9 &26.5 &  37.7 &  38.9 &  17.2 &  31.9 &  38.2 &  40.6 & \textbf{93.1} \\
        D vs W      &  52.5 &  83.1 &  88.5 &  71.9 &\textbf{89.8} &  84.7 &  87.1 &  32.5 &  69.5 &  91.5 &  87.5 & 83.1 \\\hline
        Mean        &  41.1 &  47.6  & 48.7 &  41.9 &  39.4  & 47.4 &  48.4  & 26.3  & 45.4  & 49.8  & 52.7 &  \textbf{79.3}\\
    \end{tabular}}
 \end{table}
\begin{table}
      \centering
      \caption{Mean accuracy of traditional and deep DA on Caltech-Office with Decaf and original images (Deep Learning approaches), respectively.\label{tab:result_caltech_decaf}}

    \resizebox{\textwidth}{!}{%
    \begin{tabular}{l|cccccccccccc|cccccc}
             \multicolumn{13}{c}{\textbf{Traditional Methods}} & \multicolumn{5}{c}{\textbf{Deep Domain Adaptation}} \\
\textbf{Dataset}  & SVM   & TCA  & JDA   &GFK    & SA & CORAL & CGCA & SCA &EasyTL & JGSA &MEDA & NSO (ours) & Alexnet & DDC-MMD & JAN & DAN & Deep-CORAL\\ \hline
    C vs A  &   90.6 &  90.2 & 92.4 &  85.6 &  92.0 &  91.5 &  90.1 &  48.0 &  90.2 &   92.1 &  \textbf{93.5} &  88.9   & 92.5 &  92.5 & 93.4 &  92.9 &  92.8 \\
    C vs W  &   79.0 &  78.3 & 81.7 &  76.6 &  73.2 &  78.6 &  75.9 &  35.3 &  76.9 &   86.4 &  \textbf{93.6} &  81.3   & 74.8 &  74.9 & 85.0 &  86.6 &  84.3 \\
    C vs D  &   83.4 &  89.8 & 87.3 &  82.8 &  79.0 &  84.7 &  85.4 &  46.1 &  81.5 &   92.4 &  \textbf{93.0} &  79.0   & 74.9 &  74.8 & 83.0 &  82.6 &  78.1 \\
    A vs C  &   81.9 &  81.2 & 82.7 &  76.6 &  83.8 &  83.2 &  81.6 &  43.0 &  81.7 &   85.1 & \textbf{ 87.5} &  61.6   & 85.3 &  84.9 & 84.1 &  84.1 &  80.0 \\
    A vs W  &   74.2 &  78.0 & 72.9 &  67.8 &  77.3 &  75.9 &  71.2 &  36.5 &  74.2 &   79.0 &  \textbf{88.1} &  81.2   & 65.1 &  65.2 & 85.5 &  84.5 &  84.3 \\
    A vs D  &   80.9 &  80.9 & 79.6 &  73.9 &  81.5 &  81.5 &  74.8 &  43.6 &  84.7 &   79.6 &  \textbf{91.1 }&  79.0   & 78.0 &  75.8 & 83.3 &  85.4 &  65.6 \\
    W vs C  &   63.0 &  69.5 & 74.0 &  61.1 &  76.0 &  67.9 &  73.7 &  27.9 &  66.3 &   84.9 &  \textbf{88.3} &  63.6   & 70.9 &  69.8 & 78.4 &  78.6 &  60.8 \\
    W vs A  &   73.8 &  74.6 & 79.7 &  71.2 &  86.1 &  76.0 &  80.5 &  29.8 &  73.6 &   90.3 &  93.1 &  \textbf{96.2}   & 80.0 &  77.6 & 84.5 &  83.4 &  73.6 \\
    W vs D  &  \textbf{100.0} & \textbf{100.0} &\textbf{100.0} & \textbf{100.0} &  98.7 & \textbf{100.0} &\textbf{ 100.0 }&  51.1 &  98.1 &  \textbf{100.0} & \textbf{100.0} &  79.0   & 98.8 &  98.8 & 99.7 &  99.5 &  99.4 \\
    D vs C  &   52.7 &  68.8 & 80.2 &  61.2 &  75.9 &  68.0 &  75.5 &  24.2 &  69.1 &   85.0 &  \textbf{87.1} &  66.7   & 77.3 &  77.8 & 79.6 &  78.1 &  66.5 \\
    D vs A  &   62.5 &  79.7 & 88.9 &  69.5 &  87.3 &  77.2 &  86.9 &  26.2 &  76.3 &   91.9 &  \textbf{93.2} &  92.8   & 82.8 &  82.3 & 84.4 &  85.1 &  77.4 \\
   D vs W   &   89.8 &  97.6 & 99.3 &  98.6 &  95.6 &  98.3 &  99.0 &  33.7 &  93.9 &   99.7 &  \textbf{99.0} &  83.1   & 99.0 &  98.8 & 98.7 &  98.6 &  99.0 \\\hline
       Mean &   77.7 &  82.4 & 84.9 &  77.1 &  83.9 &  81.9 &  82.9&   37.1 &   80.5&   88.9 &  \textbf{92.3} &   79.4  & 81.6 &  81.1 & 86.6 &  86.6 &  80.1 \\
    \end{tabular}}
 \end{table}
 \begin{table}
    \caption{Mean computational time in seconds of subspace DA methods. \label{tab:result_time}}
\begin{tabular}{l|cccccccccccc}
    \centering
    \textbf{Dataset}  & TCA  & JDA     & GFK & SA & CORAL & CGCA & SCA & JGSA &MEDA & NSO (ours)\\ \hline
    Newsgroup         &  21.4&   4.8   & 214.4&   59.7&  705.8  &   11977.0 & 59.0 & 3637.0 & 3447.0 & \textbf{2.64} \\
    Reuters           &  6.5 &   1.5   & 2.6 &  3.0  & 15.4 & 225.6  & 14.8 & 122.1 & 53.2  & \textbf{0.6} \\
    CO - Surf         &  3.2 &   0.9   & 0.6 &  0.7  &  0.4 &   6.4  & 12.2 &  10.8 &  6.3  & \textbf{0.2 }\\
    CO - Decaf        &  1.8 &   0.4   & 1.1 &  1.3  & 10.6 &  99.8  & 10.3 &  79.8 & 45.0  & \textbf{0.2} \\\hline
    Overall           & 8.2 &1.9 & 54.7 &16.2& 183.1& 3077.2& 24.1& 962.4 & 887.9 & 0.9\\
\end{tabular}
\end{table}
\subsection{Time Results}\label{sec:time_comparison}
The mean time results of the subspace DA methods in seconds are shown in the Tab.~\ref{tab:result_time}. The deep DA methods are not presented as they are unrivaled to the traditional methods. The experiments shows that our NSO approach is task-independent, the fastest algorithm. Compared to recent MEDA, JGSA and CGCA, the NSO approach needs substantially less time. The related SA approach is also fast, but as theoretically derived, the override of a subspace basis  approximated by Nyström leads to a boost in computational performance. In summary, the NSO approach is efficient and should be favored with regard to Green AI.

\section{Conclusion}\label{sec:conclusion}

We proposed a low-rank domain approximation algorithm called Nyström Subspace Override. It overrides the source basis with the target basis, which is designed as a domain invariant subspace projection operator. Due to the affiliation of the operator to the target space, we make sure that both domains lie in the same subspace. It requires only a subset of domain data from both domains and provides a subspace variant of the domain adaptation-related least-squares problem. The Nyström based projection, paired with smart class-wise sampling, showed its reliability and robustness in this study. Validated on common domain adaptation tasks and data, it showed a convincing performance. Additionally, NSO has the lowest computational complexity and time consumption compared to discussed solutions, which makes the approach favorable in the light of Green AI. The next steps are a theoretically evaluation of the Nyström approximation error with the proposed decomposition.
\FloatBarrier
\section*{Acknowledgment}
\scriptsize{We are thankful for support in the FuE program Informations- und Kommunikationstechnik of the StMWi, project OBerA, grant number IUK-1709-0011// IUK530/010.}

\begin{subappendices}
    \renewcommand{\thesection}{\Alph{section}}%
\section{Proof of  Subspace Override Bound}\label{sec:proof}
\begin{theorem}
    Given two rectangular matrices $\*X_t,\*X_s \in \mathbb{R}^{n \times d}$ with $n,d > 1$ and rank  of $\*X_t$ and $\*X_s > 1$. The norm $\norm{\*X_s^l-\*X_t^l}$ in the subspace $\mathbb{R}^l$ induced by normalized subspace projector $\*M\in \mathbb{R}^{n \times l}$ with $ \*M^T\*M = \*I $ is bounded by
    \begin{equation}\label{eq:norm_proof_transfer}
    E_{SO} =   \norm{\*X_s^l-\*X_t^l} < \sum_{i=1}^{l+1} (\sigma_i(\*X_s) - \sigma_i(\*X_t))^2 \leq \norm{\*X_s-\*X_t}.
    \end{equation}
\end{theorem} 
Following~\cite{horn2012matrix} the squared Frobenius norm of a matrix difference between two matrices can be bounded by
\begin{equation}\label{eq:original_bound}
    \sum_{i=1}^q (\sigma_i(\*X_s) - \sigma_i(\*X_t))^2 \leq \norm{\*X_s-\*X_t},
\end{equation}
where $q  = min(n,d)$ and $\sigma_i(\cdot)$ is the $i$-th singular value of the respective matrix in descending order.
However, the subspace matrices $\*X_s^l$ and $\*X_t^l$ are a special case due to the subspace override of the projector $\*M =\*U_t^l{\*U_s^l}^{-1} $, because
\begin{align}
    \norm{\*X_s^l-\*X_t^l} &= \norm{\*M\*U_s^l \b\Sigma_s^l - \*U_t^l \b\Sigma_t^l} =  \norm{\*U_t^l\b\Sigma_s^l - \*U_t^l\b\Sigma_t^l }  \\
    &=   \norm{\*U_t^l\b\Sigma_s^l} + \norm{\*U_t^l\b\Sigma_t^l} - 2Tr({\b\Sigma_s^l}^T {\*U_t^l}^T \*U_t^l \b\Sigma_t^l) \label{eq:bound1} \\
    &= \norm{\b\Sigma_s^l} + \norm{\b\Sigma_t^l} - 2Tr({\b\Sigma_s^l}^T \b\Sigma_t^l) \label{eq:bound2} \\
    &= \sum_{i=1}^l \sigma_i^2(\*X_s^l) +\sum_{i=1}^l \sigma_i^2(X_t^l) - 2\sum_{i=1}^l (\sigma_i(X_s^l) \cdot \sigma_i(X_t^l))\\
    &= \sum_{i=1}^l (\sigma_i(\*X_s^l) - \sigma_i(\*X_t^l))^2.
\end{align}
The important fact in the right part of Eq.~\eqref{eq:bound1} and~\eqref{eq:bound2} is that we do not rely on the bound of the Frobenious inner product as in the proof for Eq.~\eqref{eq:original_bound}~\cite[p.~459]{horn2012matrix}, because ${\*U_t^l}^T\*U_t^l = \*I$. Therefore, we can directly compute the Frobenius inner product of the the diagonal matrices $\b\Sigma_s^l$ and $\b\Sigma_t^l$, which is simply the sum of the product of the singular values. Consequently follows for $l+1$ and $(\sigma_{l+1}(\*X_s) - \sigma_{l+1}(\*X_t))^2\neq 0$,
\begin{equation}
    \norm{\*X_s^l-\*X_t^l} < \sum_{i=1}^{l+1} (\sigma_i(\*X_s) - \sigma_i(\*X_t))^2 < \sum_{i=1}^q (\sigma_i(\*X_s) - \sigma_i(\*X_t))^2 \leq \norm{\*X_s-\*X_t},
\end{equation}
where again $q  = min(n,d)$ and $1<l<q$.
\section{Mathematical Background}\label{sec:preliminaries}
We introduce the basics of the Nyström kernel approximation in Sec.~\ref{sec:nystroem_kernel}, which is the foundation of the Nyström based Singular Value Decomposition in Sec.~\ref{sec:general}. The Nyström SVD is used for constructing an approximated subspace transformation of (Nyström) Subspace Override in Sec.~\ref{sec:NSO}.
\subsection{Nyström Approximation}\label{sec:nystroem_kernel}
The computational complexity of calculating kernels or eigensystems scales with $\mathcal{O}(n^3)$ where $n$ is the sample size~\cite{NIPS2000_1866}. Therefore, low-rank approximations and dimensionality reduction of data matrices are popular methods to get better computational performance. In this scope, however, not limited to it, the Nyström approximation~\cite{NIPS2000_1866} is a reliable technique to accelerate eigendecomposition or approximation of general symmetric matrices~\cite{Gisbrecht2015}.\\
It computes an approximated set of eigenvectors and eigenvalues based on a usually much smaller sample matrix.
The landmarks are typically picked random, but advanced sampling concepts could be used as well~\cite{Kumar.2012}.
The approximation is exact if the sample size is equal to the rank of the original matrix and the rows of the sample matrix are linear independent~\cite{Gisbrecht2015}.
In general, the Nyström approximation technique assumes a symmetric matrix $\mathbf{K} \in \mathbb{R}^{n\times n}$ with a decomposition of the form
\begin{equation}\label{eq:matrix_decomposition}
    \mathbf{K}=
    \begin{bmatrix}
        \mathbf{A} & \mathbf{B} \\
        \mathbf{C} & \mathbf{F} \\
    \end{bmatrix},
\end{equation}
with $\*{A} \in \mathbb{R}^{l\times l}$, $\*{B} \in \mathbb{R}^{l\times (n-l)}$, $\*{C} \in \mathbb{R}^{(n-l)\times l} $ and $\*{F} \in \mathbb{R}^{(n-l)\times (n-l)}$.
The matrix $\*{A}$ is called the landmark matrix containing $l$ randomly chosen rows and columns from $\*K$ and has the \ac{EVD} $\*{A}=\*{U}\boldsymbol{\Lambda}\*{U}^{-1}$.
The eigenvectors are $\*{U}\in \mathbb{R}^{l\times l}$ and the eigenvalues are on the diagonal of $\boldsymbol{\Lambda}\in \mathbb{R}^{l\times l}$.
The remaining approximated eigenvectors $\hat{\*U}$ of $\*K$ as part $\*C$ or $\*B^T$, are obtained by the Nyström method with $\hat{\*{U}}\boldsymbol{\Lambda}=\*{CU}$.
Combining $\*U$ and $\hat{\*U}$ the \textit{full} approximated eigenvectors of $\*{K}$ are
\begin{equation}\label{eq:ny_eigenvector}
    \tilde{\*{U}} =
        \begin{bmatrix}
        \*U \\
        \hat{\*U}
    \end{bmatrix}
    =
    \begin{bmatrix}
        \*U \\
        \*C \*U \boldsymbol{\Lambda}^{-1}
    \end{bmatrix} \in \mathbb{R}^{n\times l}.
\end{equation}
The right part of the EVD ($\tilde{\*U}^{-1}$) of $\*K$ can be obtained via Nyström similar to~\eqref{eq:ny_eigenvector} by
\begin{equation}\label{eq:ny_left_eigenvector}
    \tilde{\mathbf{V}}
    =\begin{bmatrix}
        \mathbf{U}^{-1}&&\boldsymbol{\Lambda}^{-1}\mathbf{U}^{-1}\mathbf{B}
    \end{bmatrix}.
\end{equation}
Combining~\eqref{eq:ny_eigenvector}, ~\eqref{eq:ny_left_eigenvector} and $\boldsymbol{\Lambda}$, the matrix $\mathbf{K}$ is approximated by
\begin{equation}\label{eq:ny_decomposition}
        \tilde{\mathbf{K}}= \tilde{\mathbf{U}} \boldsymbol{\Lambda} \tilde{\mathbf{V}}=
        \begin{bmatrix}
            \mathbf{U} \\
            \mathbf{CU}\boldsymbol{\Lambda}^{-1}
        \end{bmatrix}
        \boldsymbol{\Lambda}
        \begin{bmatrix}
            \mathbf{U}^{-1}& \boldsymbol{\Lambda}^{-1}\mathbf{U}^{-1}\mathbf{B}
        \end{bmatrix}.
\end{equation}
The Frobenius Norm gives the Nyström approximation error between ground truth and reconstructed matrices, i.\,e. $E_{ny}=||\tilde{\mathbf{K}} - \mathbf{K}||_F$, with bounds proven by~\cite{GittensM16}.

\subsection{General Matrix Approximation}\label{sec:general}
Another application of the Nyström method is the approximation of the Singular Value Decomposition, which generalizes the concept of matrix decomposition with the consequence that respective matrices must not be squared~\cite{Nemtsov2016}.\\
Let $\*X \in \mathbb{R}^{n\times d}$ be a rectangular data matrix. Following~\cite{Nemtsov2016}, a decomposition as in~\eqref{eq:matrix_decomposition} can be obtained.
The SVD of the landmark matrix is given by $\*A = \*U \b\Sigma \*V^T$ where $\*U$ are left, and $\*V$ are right singular vectors.
$\b\Sigma$ are non-negative singular values.
The left and right singular vectors for the non-symmetric part $\*C$ and $\*B$ are obtained via Nyström techniques and are defined as $\hat{\*U} =  \*C \*V \b\Sigma^{-1}$ and $\hat{\*V} = \*B^T \*U \b\Sigma^{-1} $ respectively~\cite{Nemtsov2016}.
Applying the same principal as for Nyström-\ac{EVD}, $\*X$ is approximated by
\begin{align}\label{eq:ny_svd}
    \tilde{\*X} = \tilde{\*U} \b\Sigma \tilde{\*V}^T &=
    \begin{bmatrix}
        \*U \\
        \hat{\*U}
    \end{bmatrix}
    \b\Sigma
    \begin{bmatrix}
        \*V &  \hat{\*V}
    \end{bmatrix}\\
    &=
    \begin{bmatrix}
        \*U \\
        \*C \*V \b\Sigma^-1
    \end{bmatrix}
    \b\Sigma
    \begin{bmatrix}
        \*V &  \b\Sigma^{-1} \*U^T \*B
    \end{bmatrix}.
\end{align}
\subsection{Gerschgorin Theorem}\label{app:gerschgorin}
The Gerschgorin theorem~\cite{varga2010} provides a geometric structure to bound eigenvalues to so-called discs for complex square matrices, but also generalize to none complex square matrices. The work of~\cite{qi1984some} expands the Gerschgorin circles to so-called Gerschgorin type circles for singular values:
\begin{theorem}[Gerschgorin Type Bound for Singular Values~\cite{qi1984some}]\label{theo:discs}
    Given the matrix $\*X\in \mathbb{R}^{n\times d}$ with $n,d \geq 1$, the singular values $\{s_i\}_{i=1}^n$ of $\*X$ are in the range of
    \begin{equation}
        s_i = \{ p_i \pm |r_i|\}, \>\>\>\>i=1,\dots,n.
    \end{equation}
     Where $p_i = |x_{ii}|$, $x \in \*X$ and the range $r_i$ is defined as
    \begin{equation}
        r_i = \max \Biggl\{\sum_{j=1,j\neq i}^{d} |x_{ij}|, \sum_{j=1,j\neq i }^n |x_{ji}| \Biggl\}, \>\>\>\>i=1,\dots,n.
    \end{equation}
\end{theorem}
By using theorem~\ref{theo:discs} we can bound the norm of the singular values $\norm{\b\Sigma}$ of $\*X$ by the square root of the squared sum of the numerical range given by
\begin{equation}\label{eq:bound_singular_values}
    D = \norm{\b\Sigma} \leq \sqrt{\sum_{i=1}^n (p_i+|r_i|)^2}.
\end{equation}
\section{Component Analysis}\label{sec:add_results}
We inspect the performance contribution of the different parts of the NSO approach. First, the exact solution to the optimization problem is called Subspace Override (SO). The approximation with uniform sampling is evaluated to study the impact of class-wise sampling on the performance. To show the efficiency of the subspace projection in original space, we include a kernelized version where we approximate the RBF-kernels of $\*X_s$ and $\*X_t$, respectively.
The results are given in Tab.~\ref{tab:components} and show that the Nyström approximation independent of the sampling strategy yields the best performance. This comes from the approximation of the subspace projection, where small values are likely to be zero, hence reducing noise further. The kernelized version is not recommended due to bad performance. Overall, as proposed, the class-wise NSO is recommended, because it is slightly better.
\begin{table}%
\centering
\caption{Component evaluation of NSO in mean accuracy.\label{tab:components}}
\begin{tabular}{l|cccccccHHcc}
   \textbf{Dataset}          & \textbf{SO}  & \textbf{NSO$_{uniform}$}   & \textbf{NSO$_{classwise}$}    & \textbf{NSO$_{ker}$} \\\hline
   Reuters     &  94.8  &   \textbf{97.6}  &    \textbf{97.6 }   &  80.8 \\
   Newsgroup   &  93.0  &    96.1  &    \textbf{97.4}    &  94.3 \\
   CO - Surf   &  \textbf{79.3}  &    79.1  &    \textbf{79.3}    &  56.5 \\
   CO - Decaf  &  79.2  &    \textbf{79.4}  &    \textbf{79.4}    &  76.4 \\\hline
   Overall     &  86.2  &    \textbf{88.1}  &    \textbf{88.4}    &  77.0 \\
\end{tabular}
\end{table}
\end{subappendices}
\bibliographystyle{splncs04}
\bibliography{raab_schleif_nso}

\begin{thebibliography}{10}
\providecommand{\url}[1]{\texttt{#1}}
\providecommand{\urlprefix}{URL }
\providecommand{\doi}[1]{https://doi.org/#1}

\bibitem{Aljundi2015}
Aljundi, R., Emonet, R., Muselet, D., Sebban, M.: {Landmarks-based kernelized
  subspace alignment for unsupervised domain adaptation}. In: 2015 IEEE
  Conference on Computer Vision and Pattern Recognition (CVPR). vol.
  07-12-June, pp. 56--63. IEEE (jun 2015)

\bibitem{Blitzer2011}
Blitzer, J., Foster, D., Kakade, S.: {Domain adaptation with coupled
  subspaces}. Journal of Machine Learning Research  \textbf{15},  173--181
  (2011)

\bibitem{Dai2007}
Dai, W., Yang, Q., Xue, G.R., Yu, Y.: {Boosting for transfer learning}. In:
  Proceedings of the 24th international conference on Machine learning - ICML
  '07. pp. 193--200. ACM Press, New York, New York, USA (2007)

\bibitem{Donahue2014}
Donahue, J., Jia, Y., Vinyals, O., Hoffman, J., Zhang, N., Tzeng, E., Darrell,
  T.: {DeCAF: A deep convolutional activation feature for generic visual
  recognition}. 31st International Conference on Machine Learning, ICML 2014
  \textbf{2},  988--996 (2014)

\bibitem{Elhadji-Ille-Gado2018}
Elhadji-Ille-Gado, N., Grall-Maes, E., Kharouf, M.: {Transfer Learning for
  Large Scale Data Using Subspace Alignment}. In: 2017 16th IEEE International
  Conference on Machine Learning and Applications (ICMLA). vol. 2018-Janua, pp.
  1006--1010. IEEE (dec 2017)

\bibitem{Fernando2013a}
Fernando, B., Habrard, A., Sebban, M., Tuytelaars, T.: {Unsupervised visual
  domain adaptation using subspace alignment}. Proceedings of the IEEE
  International Conference on Computer Vision pp. 2960--2967 (2013)

\bibitem{Ghifary2017}
Ghifary, M., Balduzzi, D., Kleijn, W.B., Zhang, M.: {Scatter Component
  Analysis: A Unified Framework for Domain Adaptation and Domain
  Generalization}. IEEE Transactions on Pattern Analysis and Machine
  Intelligence  \textbf{39}(7),  1414--1430 (jul 2017)

\bibitem{Gisbrecht2015}
Gisbrecht, A., Schleif, F.M.: {Metric and non-metric proximity transformations
  at linear costs}. Neurocomputing  \textbf{167},  643--657 (2015)

\bibitem{GittensM16}
Gittens, A., Mahoney, M.W.: {Revisiting the Nystrom Method for Improved
  Large-Scale Machine Learning}. Journal of Machine Learning Research
  \textbf{17},  117:1----117:65 (2013)

\bibitem{Gong2012}
Gong, B., Shi, Y., Sha, F., Grauman, K.: {Geodesic flow kernel for unsupervised
  domain adaptation}. In: Proceedings of the IEEE Computer Society Conference
  on Computer Vision and Pattern Recognition. pp. 2066--2073 (2012)

\bibitem{horn2012matrix}
Horn, R.A., Johnson, C.R.: {Matrix analysis}. Cambridge university press (2012)

\bibitem{Krizhevsky2012}
Krizhevsky, A., Sutskever, I., Hinton, G.E.: {ImageNet Classification with Deep
  Convolutional Neural Networks}. In: Pereira, F., Burges, C.J.C., Bottou, L.,
  Weinberger, K.Q. (eds.) Advances in Neural Information Processing Systems 25,
  pp. 1097--1105. Curran Associates, Inc. (2012)

\bibitem{7486497}
Liu, P., Yang, P., Huang, K., Tan, T.: {Uniform low-rank representation for
  unsupervised visual domain adaptation}. In: 2015 3rd IAPR Asian Conference on
  Pattern Recognition (ACPR). pp. 216--220 (nov 2015)

\bibitem{Long2018}
Long, M., Cao, Y., Cao, Z., Wang, J., Jordan, M.I.: {Transferable
  Representation Learning with Deep Adaptation Networks}. IEEE Transactions on
  Pattern Analysis and Machine Intelligence  \textbf{PP}(c), ~1 (2018)

\bibitem{Long2014a}
Long, M., Wang, J., Ding, G., Pan, S.J., Yu, P.S.: {Adaptation regularization:
  A general framework for transfer learning}. IEEE Transactions on Knowledge
  and Data Engineering  \textbf{26}(5),  1076--1089 (2014)

\bibitem{Long2013a}
Long, M., Wang, J., Ding, G., Sun, J., Yu, P.S.: {Transfer feature learning
  with joint distribution adaptation}. Proceedings of the IEEE International
  Conference on Computer Vision pp. 2200--2207 (2013)

\bibitem{Long2015}
Long, M., Wang, J., Sun, J., Yu, P.S.: {Domain invariant transfer kernel
  learning}. IEEE Transactions on Knowledge and Data Engineering
  \textbf{27}(6),  1519--1532 (2015)

\bibitem{LongJAN2017}
Long, M., Zhu, H., Wang, J., Jordan, M.I.: {Deep Transfer Learning with Joint
  Adaptation Networks}. In: Proceedings of the 34th International Conference on
  Machine Learning - Volume 70. pp. 2208--2217. ICML'17, JMLR.org (2017)

\bibitem{Mahadevan19}
Mahadevan, S., Mishra, B., Ghosh, S.: {A Unified Framework for Domain
  Adaptation Using Metric Learning on Manifolds}. In: Berlingerio, M., Bonchi,
  F., G{\"{a}}rtner, T., Hurley, N., Ifrim, G. (eds.) Machine Learning and
  Knowledge Discovery in Databases. pp. 1--17. Springer International
  Publishing, Cham (2019)

\bibitem{Nemtsov2016}
Nemtsov, A., Averbuch, A., Schclar, A.: {Matrix compression using the
  Nystr{\"{o}}m method}. Intelligent Data Analysis  \textbf{20}(5),  997--1019
  (may 2016)

\bibitem{Pan2011}
Pan, S.J., Tsang, I.W., Kwok, J.T., Yang, Q.: {Domain adaptation via transfer
  component analysis}. IEEE Transactions on Neural Networks  \textbf{22}(2),
  199--210 (2011)

\bibitem{5288526}
Pan, S.J., Yang, Q.: {A survey on transfer learning}. IEEE Transactions on
  Knowledge and Data Engineering  \textbf{22}(10),  1345--1359 (2010)

\bibitem{qi1984some}
Qi, L.: {Some simple estimates for singular values of a matrix}. Linear Algebra
  and Its Applications  \textbf{56},  105--119 (1984)

\bibitem{stvm}
Raab, C., Schleif, F.M.: {Sparse transfer classification for text documents}.
  In: Trollmann, F., Turhan, A.Y. (eds.) Lecture Notes in Computer Science
  (including subseries Lecture Notes in Artificial Intelligence and Lecture
  Notes in Bioinformatics). vol. 11117 LNAI, pp. 169--181. Springer
  International Publishing (2018)

\bibitem{DBLP:journals/ijon/SchleifGT18}
Schleif, F., Gisbrecht, A., Ti{\~{n}}o, P.: Supervised low rank indefinite
  kernel approximation using minimum enclosing balls. Neurocomputing
  \textbf{318},  213--226 (2018)

\bibitem{Shao2018}
Shao, J., Huang, F., Yang, Q., Luo, G.: {Robust Prototype-Based Learning on
  Data Streams}. IEEE Transactions on Knowledge and Data Engineering
  \textbf{30}(5),  978--991 (2018)

\bibitem{Shao2014a}
Shao, M., Kit, D., Fu, Y.: {Generalized transfer subspace learning through
  low-rank constraint}. International Journal of Computer Vision
  \textbf{109}(1-2),  74--93 (2014)

\bibitem{sun2016return}
Sun, B., Feng, J., Saenko, K.: {Return of Frustratingly Easy Domain
  Adaptation}. In: Proceedings of the Thirtieth {AAAI} Conference on Artificial
  Intelligence, February 12-17, 2016, Phoenix, Arizona, {USA.} pp. 2058--2065
  (2016)

\bibitem{Baochen2016}
Sun, B., Saenko, K.: {Deep CORAL: Correlation Alignment for Deep Domain
  Adaptation}. In: Hua, G., J{\'{e}}gou, H. (eds.) Computer Vision -- ECCV 2016
  Workshops. pp. 443--450. Springer International Publishing, Cham (2016)

\bibitem{Kumar.2012}
Talwalkar, A., Kumar, S., Mohri, M.: {Sampling Methods for the Nystr{\"{o}}m
  Method}. Journal of Machine Learning Research  \textbf{13},  981--1006 (2012)

\bibitem{Tzeng2014}
Tzeng, E., Hoffman, J., Zhang, N., Saenko, K., Darrell, T.: {Deep Domain
  Confusion: Maximizing for Domain Invariance}. CoRR  \textbf{abs/1412.3}
  (2014)

\bibitem{varga2010}
Varga, R.S.: {Ger{\v{s}}gorin and His Circles}, Springer Series in
  Computational Mathematics, vol.~36. Springer Berlin Heidelberg, Berlin,
  Heidelberg (2004)

\bibitem{WangEasyTL}
Wang, J., Chen, Y., Yu, H., Huang, M., Yang, Q.: {Easy Transfer Learning By
  Exploiting Intra-Domain Structures}. In: 2019 IEEE International Conference
  on Multimedia and Expo (ICME). pp. 1210--1215. IEEE (jul 2019)

\bibitem{meda2018}
Wang, J., Feng, W., Chen, Y., Yu, H., Huang, M., Yu, P.S.: {Visual Domain
  Adaptation with Manifold Embedded Distribution Alignment}. In: 2018 ACM
  Multimedia Conference on Multimedia Conference - MM '18. pp. 402--410. MM
  '18, ACM Press, New York, New York, USA (2018)

\bibitem{Weiss2016}
Weiss, K., Khoshgoftaar, T.M., Wang, D.: {A survey of transfer learning}.
  Journal of Big Data  \textbf{3}(1), ~9 (2016)

\bibitem{NIPS2000_1866}
Williams, C., Seeger, M.W.: {Using the Nystrom Method to Speed Up Kernel
  Machines}. In: Leen, T.K., Dietterich, T.G., Tresp, V. (eds.) NIPS
  Proceedings, vol.~13, pp. 682--688. MIT Press (2001)

\bibitem{Xiao2019}
Xiao, T., Liu, P., Zhao, W., Liu, H., Tang, X.: {Structure preservation and
  distribution alignment in discriminative transfer subspace learning}.
  Neurocomputing  \textbf{337},  218--234 (apr 2019)

\bibitem{8100030}
Zhang, J., Li, W., Ogunbona, P.: {Joint Geometrical and Statistical Alignment
  for Visual Domain Adaptation}. In: 2017 IEEE Conference on Computer Vision
  and Pattern Recognition (CVPR). pp. 5150--5158. IEEE (jul 2017)

\end{thebibliography}
\end{document}